\documentclass[pdflatex,sn-nature, table, dvipsnames, title, toc, page, figurewithintext]{sn-jnl}
\usepackage{graphicx} 
\usepackage{amsmath}
\usepackage{amssymb}
\usepackage{subcaption} 
\usepackage{tikz} 
\usepackage[percent]{overpic}
\usepackage{placeins}

\definecolor{kigaliHotspot}{HTML}{ffcf4b}
\definecolor{kigaliColdspot}{HTML}{4bfff0}

\usepackage{svg} 
\usepackage{comment}
\usepackage{makecell}
\usepackage{booktabs}
\usepackage{gensymb}
\usepackage[numbers]{natbib}
\usepackage{nomencl}
\makenomenclature
\usepackage{hyperref}
\usepackage{siunitx}
\usepackage{colortbl} 
\begin{document}

\begin{titlepage}
    \centering
    \vspace*{2cm}
    
    {\LARGE\bfseries Detecting Urban $\text{PM}_{2.5}$ Hotspots with Mobile Sensing and Gaussian Process Regression \par}
    \vspace{2cm}
    
    {\large Niál Perry\textsuperscript{a,1}, Peter P. Pedersen\textsuperscript{a,b}, Charles N. Christensen\textsuperscript{b},\\ Emanuel Nussli\textsuperscript{a}, Sanelma Heinonen\textsuperscript{a}, Lorena Gordillo Dagallier\textsuperscript{b},\\ Raphaël Jacquat\textsuperscript{b}, Sebastian Horstmann\textsuperscript{b},
    Christoph Franck\textsuperscript{b}
    \par}
    
    \vspace{1.5cm}
    
    {\textsuperscript{a} Eidgenössische Technische Hochschule Zürich, Rämistrasse 101, Zürich, Switzerland, 8092 \par}
    {\textsuperscript{b} University of Cambridge, Philippa Fawcett Drive, Cambridge, CB3 0AS\par}
    {\textsuperscript{1} Corresponding author: nial@perry-online.co.uk \par}

    \vfill
    
    {\today}
\end{titlepage}

\section*{Abstract}

Low-cost mobile sensors can be used to collect $\text{PM}_{2.5}$ concentration data throughout an entire city. However, identifying air pollution hotspots from the data is challenging due to the uneven spatial sampling, temporal variations in the background air quality, and the dynamism of urban air pollution sources. This study proposes a method to identify urban $\text{PM}_{2.5}$ hotspots that addresses these challenges, involving four steps: (1) equip citizen scientists with mobile $\text{PM}_{2.5}$ sensors while they travel; (2) normalise the raw data to remove the influence of background ambient pollution levels; (3) fit a Gaussian process regression model to the normalised data and (4) calculate a grid of spatially explicit ‘hotspot scores’ using the probabilistic framework of Gaussian processes, which conveniently summarise the relative pollution levels throughout the city. We apply our method to create the first ever map of $\text{PM}_{2.5}$ pollution in Kigali, Rwanda, at a 200m resolution. Our results suggest that the level of ambient $\text{PM}_{2.5}$ pollution in Kigali is dangerously high, and we identify the hotspots in Kigali where pollution consistently exceeds the city-wide average. We also evaluate our method using simulated mobile sensing data for Beijing, China, where we find that the hotspot scores are probabilistically well calibrated and accurately reflect the ‘ground truth’ spatial profile of $\text{PM}_{2.5}$ pollution. Thanks to the use of open-source software, our method can be re-applied in cities throughout the world with a handful of low-cost sensors. The method can help fill the gap in urban air quality information and empower public health officials.

\newpage

\renewcommand{\arraystretch}{0.9}

\section*{Nomenclature}
\begin{tabular}{@{}ll@{}} 
{$x_w$}              & Window-length parameter for computing the temporal baseline \\
{$I$}                & Index set for mobile sensing observations \\
{$i \in I$}          & Index for a mobile sensing observation \\
{$\mathbf{x}_i$}     & Latitude and longitude where $i$'th observation is recorded \\
{$y_i^{\text{raw}}$} & Raw (observed) $\text{PM}_{2.5}$ concentration for the $i$'th observation \\
{$y_i$}              & Background-normalised $\text{PM}_{2.5}$ concentration for the $i$'th observation \\
{f($\mathbf{x}_i$)}  & Random variable for the normalised $\text{PM}_{2.5}$ at location $\mathbf{x}_i$ \\
{$\varepsilon_i$}    & Measurement error associated with observation $i$ \\
{$\text{median}(y)$} & Median of $\{y_i : i \in I\}$ \\
{$J$}                & Index set of the tiles \\
{$j \in J$}          & Index for a tile \\
{$\mathcal{X}$}      & Latitude, longitude extent of the city \\
{$\mathcal{Y}$}      & Range of the normalised $\text{PM}_{2.5}$ measurements \\
{$\mathbf{x}_j$}     & Latitude and longitude of the centroid of the $j$'th tile \\
{$h(j)$}             & Hotspot score of tile $j$ \\
{$h(\mathbf{x}_i)$}  & Hotspot score of tile containing $\mathbf{x}_i$ \\
{$t_i$}              & Timestamp of the $i$'th observation \\
{$hm_t$}             & Hourly multiplier at time $t$ \\
{$\kappa$}           & Bernoulli random variable \\
{$\gamma$}           & Gamma random variable representing random pollution spikes \\
{$\text{PM}^j_{2.5}$}& Average $\text{PM}_{2.5}$ value in tile $j$ (for simulating the data) \\
{$R(v)$}             & Rank vector of the elements of some vector $v$ \\
\end{tabular}


\newpage

\section{Introduction}

Air pollution is considered `the single biggest environmental threat to human health' by the World Health Organisation \citep{WHO2021}. A particularly harmful category of pollutants is $\text{PM}_{2.5}$, defined as particulate matter with an aerodynamic diameter $\leq 2.5$\si{\micro\meter}. $\text{PM}_{2.5}$ particulates are small enough to penetrate deep into the lungs and sometimes even the bloodstream, and are associated with a wide range of adverse respiratory \citep{Xing2016-cr}, cardiovascular \citep{ZHANG2023162191}, and cerebrovascular health outcomes \citep{Gutierrez-Avila2018-uh}. Increased exposure to ambient $\text{PM}_{2.5}$ has also been linked to adverse birth outcomes \citep{Stieb2016-cv} and increased incidence of diabetes mellitus \citep{Bowe2018-ri}. An estimated 4.1 million deaths in 2019 can be attributed to long-term ambient $\text{PM}_{2.5}$ exposure \citep{GBD_2019_Diseases_and_Injuries_Collaborators2020-je}, and it is estimated that PM2·5-attributable deaths increased in all
global regions except Europe and the Americas from 2000-2019 \citep{Southerland2022-gn}, a concerning trend. 

In 2015, the 193 member states of the United Nations ratified the Sustainable Development Goals (SDGs), including SDG 11.6 which seeks to improve urban air quality by 2030. Despite receiving global support, current $\text{PM}_{2.5}$ monitoring infrastructure is insufficient to measure whether SDG 11.6 is achieved. As of 2019, the majority (141) of the world's countries had no regular $\text{PM}_{2.5}$ monitoring at all, and the global mean population distance to a $\text{PM}_{2.5}$ monitor was 220km \citep{Martin2019-jj}. A more recent analysis estimates that in-situ $\text{PM}_{2.5}$ measurements are missing for over 50\% of the world's urban population \citep{Apte2021-eh}. $\text{PM}_{2.5}$ estimates based on satellite data are a welcome contribution to filling the data gap (e.g., \cite{Wei2023-sv}). However, satellite estimates often have a granularity of hundreds of meters or even kilometers, which is too coarse to capture the full variability of $\text{PM}_{2.5}$ within a city and accurately assess citizens' personal exposure. Furthermore, satellite estimates can exhibit errors in the range 22-85\% when they are not calibrated by ground $\text{PM}_{2.5}$ measurements \citep{Alvarado2019-oa}. 

Low-cost sensors are a promising alternative to satellite data, and several studies have already explored using the technology to estimate the concentration of $\text{PM}_{2.5}$ throughout an urban area. Low-cost sensors can be installed at fixed sites to increase the spatial coverage of the city's monitoring network \citep{Gao2015, Lu2021, Considine2021}. An alternative approach that more easily scales for city-wide data capture is mobile sensing, wherein the sensors are transported while they record geolocated air quality measurements. Several studies have deployed mobile sensors along predetermined routes on multiple different days and/or at different times of the day \citep{Cummings2021-fw, Hart2020-aq, Lim2019-ji}. This approach allows one to analyse how pollution levels vary along a fixed route, as well as how pollution varies across time when the location is controlled. However, deploying sensors along prescribed routes is not optimal for assessing pollution exposure because the routes may differ from routes that people usually take, and the spatial coverage is often limited. Other studies have instead fit mobile sensors to people \citep{Bousiotis2024-pu} or vehicles such as taxis \citep{Leung2019-dd, anastasios2025} or trams \citep{Li2021-tc} while going about their daily business. This leads to $\text{PM}_{2.5}$ concentrations data that are more informative about people's day-to-day exposure and the data will often cover a larger spatial area than predetermined routes. 

Once pollutant concentrations have been recorded with mobile sensing, a key question is where the air pollution hotspots are in the city. A hotspot is a location where the average pollution level exceeds that of the surrounding area. Identifying hotspots is relevant from a health perspective because short-term exposure to high levels of air pollution has been positively associated with increased inflammatory markers in the lungs \citep{Dauchet2018-eg} and with increased all-cause mortality for the pollutants $\text{PM}_{2.5}$, PM10, NO2 and O3 \citep{Orellano2020-lz}. The evidence suggests that health outcomes are not just determined by average $\text{PM}_{2.5}$ exposure, but also by the maximum or near-maximum $\text{PM}_{2.5}$ exposure. Additionally, hotspots are relevant from a policymaking perspective, as they constitute locations where policymakers can focus their clean air initiatives to achieve the maximal improvement in health outcomes. Hotspot detection can also help to pinpoint $\text{PM}_{2.5}$ pollution sources that were previously unknown  \citep{Collado2023}.

In this study we propose an urban $\text{PM}_{2.5}$ hotspot-detection method that combines mobile sensing data with Gaussian process regression. Gaussian process regression has three particularly attractive properties for modelling $\text{PM}_{2.5}$ concentrations data: (1) it allows for a rich specification of the target function \citep{Gelfand2016-qt}; (2) it is flexible, accommodating diverse data types and supplementary data (e.g., meteorological variables); (3) compared to many geospatial models, specifying a Gaussian process requires little expert intervention \citep{Christianson2023-rp, Oliver2014-qb}. These advantageous properties have motivated increased research into applying Gaussian process regression to predict $\text{PM}_{2.5}$ concentrations \citep{Stoddart2023-ug, Liu2018-dg, Wang2021-lg} and to interpolate sparse $\text{PM}_{2.5}$ concentrations data \citep{Tang2017-pe, Cheng2014-fm}. However, to the best of our knowledge, this study is the first to apply Gaussian process regression with the explicit aim of $\text{PM}_{2.5}$ hotspot detection from mobile sensing data.

Our method contributes to the growing body of literature on identifying urban $\text{PM}_{2.5}$ hotspots. One study \cite{Goyal2021-oh} considers data from stationary $\text{PM}_{2.5}$ sensors and identify hotspots based on the frequency, scale, and consistency with which pollution levels exceed a given threshold (see also \cite{Bhardwaj2024-fm}). However, their approach is designed for fixed $\text{PM}_{2.5}$ monitoring stations rather than for the uneven sampling pattern and wide spatial coverage of mobile sensing. Another approach which is more compatible with mobile sensing considers a `window' around every observation, consisting either of neighbouring observations \citep{Zhang2021-xz} or neighbouring pixels of gridded observations \citep{Zheng2021-gf}, and identifies $\text{PM}_{2.5}$ spikes within these windows as hotspots. Although effective for localized hotspot detection, these window-based approaches lack a shared baseline across windows, making them less suitable for city-wide hotspot identification. The Gi* statistic \citep{Getis1992-cc} is another popular choice for $\text{PM}_{2.5}$ hotspot identification \citep{Cummings2021-fw, Ruidas2022-jb}. A location is considered a hotspot if its Gi* statistic value is significantly larger than one would expect under a spatially-random distribution of the $\text{PM}_{2.5}$ observations. A limitation is that the Gi* are merely descriptive statistics about the clustering of observed data, and they convey less information than probabilistic approaches which also report the variability in the observed data.

\begin{figure}
    \centering
    \begin{minipage}[b]{0.48\linewidth}
        \centering
        \includegraphics[width=\linewidth]{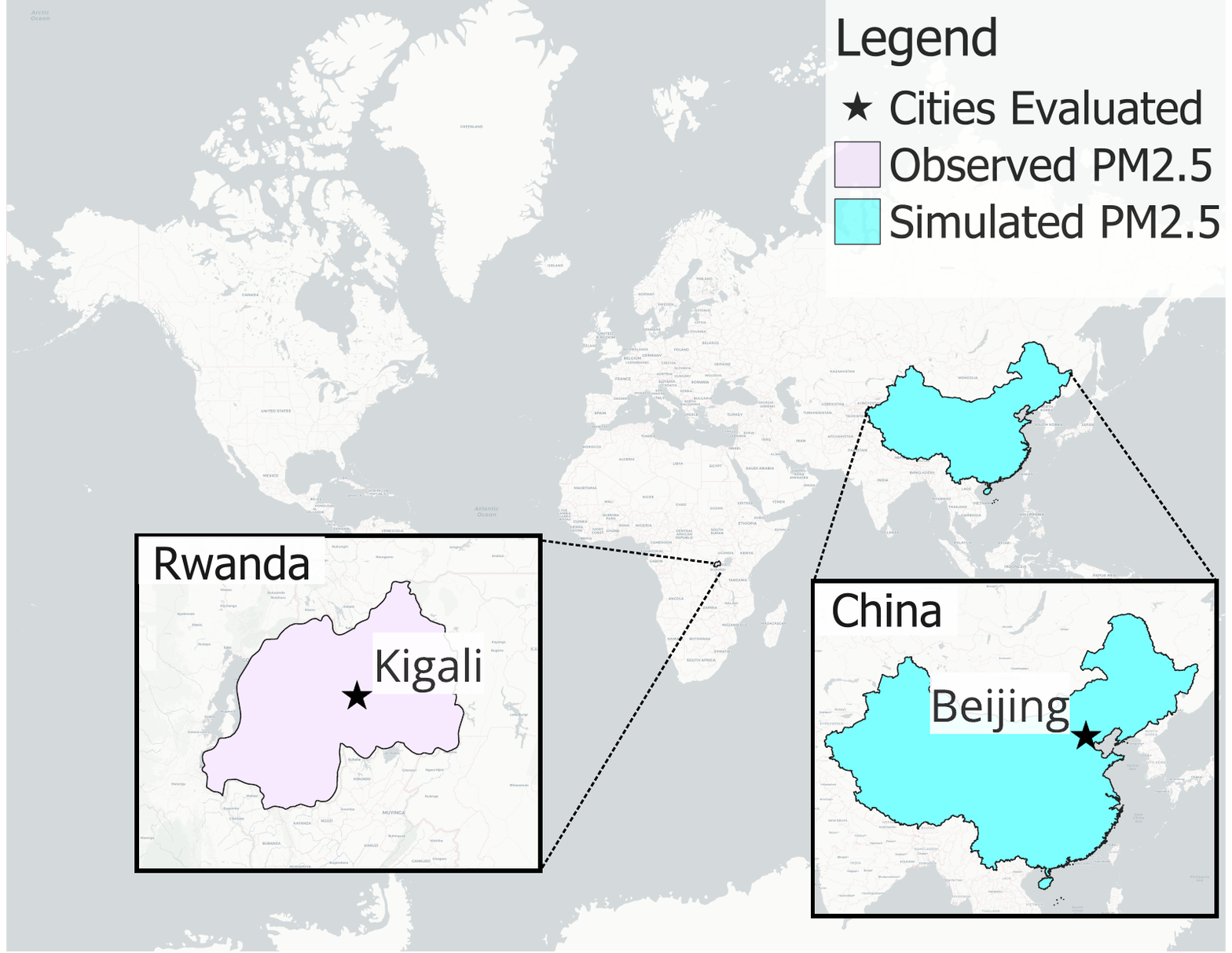}
        \subcaption{City locations.}
    \end{minipage}
    \hfill
    \begin{minipage}[b]{0.48\linewidth}
        \centering
        \renewcommand{\arraystretch}{1.4} 
        \normalsize
        \resizebox{!}{\height}{ 
        \begin{tabular}{|p{2cm}|p{3cm}|p{3cm}|}
            \hline
            City & \cellcolor[HTML]{F6EAFF}Kigali & \cellcolor[HTML]{9BFFFE}Beijing \\
            \hline
            Country & \cellcolor[HTML]{F6EAFF}Rwanda & \cellcolor[HTML]{9BFFFE}China \\
            \hline
            Coordinates & \cellcolor[HTML]{F6EAFF}1\degree56\textquotesingle38\textquotedblright S\quad30\degree3\textquotesingle34\textquotedblright E & \cellcolor[HTML]{9BFFFE}39\degree54\textquotesingle24\textquotedblright N\quad116\degree23\textquotesingle51\textquotedblright E \\
            \hline
            Mobile \newline sensing & \cellcolor[HTML]{F6EAFF}Observed & \cellcolor[HTML]{9BFFFE}Simulated \\
            \hline
            Spatial data source & \cellcolor[HTML]{F6EAFF}N/A & \cellcolor[HTML]{9BFFFE}\cite{Wang2023-fa} \\
            \hline
            Temporal data source & \cellcolor[HTML]{F6EAFF}N/A & \cellcolor[HTML]{9BFFFE}US Embassy \newline Beijing \\
            \hline
        \end{tabular}
        }
        \subcaption{Sources of data.}
    \end{minipage}
    \caption{Datasets on which the method is evaluated.}
    \label{fig:countries_plot}
\end{figure}

To address these limitations, we propose a Gaussian process-based approach that is compatible with spatially- and temporally-uneven mobile sensing data, identifies hotspots relative to the whole city, and outputs spatially-explicit hotspot scores that capture the variability in the observed data. We proceed to evaluate the $\text{PM}_{2.5}$ hotspot-detection method on two datasets (Figure \ref{fig:countries_plot}). The first is a mobile sensing dataset that we gathered in Kigali, Rwanda, in September 2021. The population of Kigali has grown from tens of thousands of people to over 1 million in the last 50 years, yet as of 2021 the city had no publicly-accessible reference $\text{PM}_{2.5}$ monitoring station. Kigali is therefore a good representation of a city in a low-income country with an air quality data gap that mobile sensing can help to resolve, in pursuit of SDG 11.6. The second dataset comprises seasonal estimates of the average $\text{PM}_{2.5}$ concentrations for Beijing. Published by \cite{Wang2023-fa}, the dataset is based on several information sources, such as satellite top-of-atmosphere data, traffic emissions and land use data, and contains average $\text{PM}_{2.5}$ estimates at a spatial resolution of 30 meters. Treating the Beijing dataset as a `ground truth' for hotspot locations, we first simulate mobile sensing data, and then evaluate whether our method can successfully reconstruct the spatial $\text{PM}_{2.5}$ distribution from the simulated mobile sensing data. 

\section{Materials and Methods}

\subsection{Mobile Sensors}

\begin{figure}
    \centering
    \includegraphics[width=0.4\linewidth]{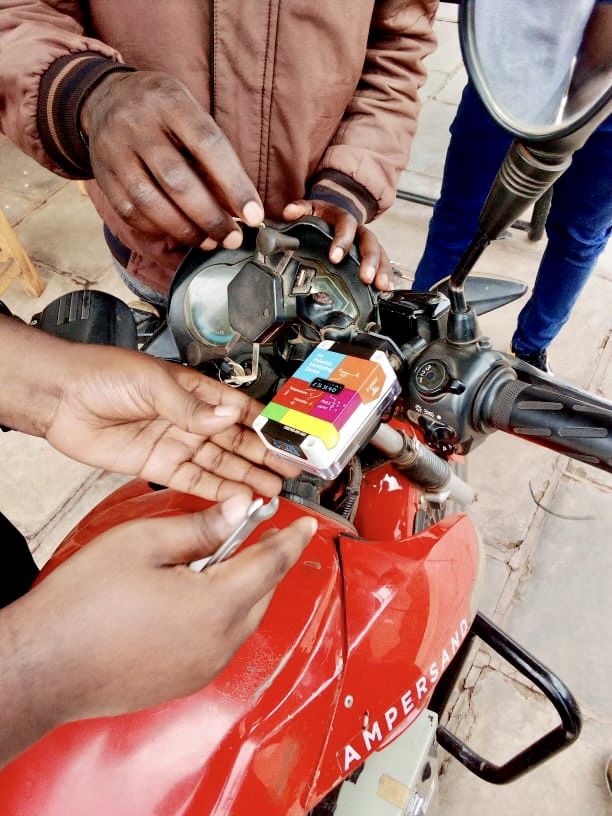}
    \caption{The open-seneca air quality monitor mounted on an electric motorbike, part of Ampersand's delivery fleet. The monitor was a portable device that measured $\text{PM}_{2.5}$, temperature, humidity, and GPS location at 1~Hz. It featured an onboard display and a Garmin-style mount for wearable or vehicle use. Raw data were stored locally on an SD card for manual upload. For the period of data collection in Kigali, the sensors were permanently installed and powered by the electric motorcycle’s onboard battery.}
    \label{fig: ampersand-sensor-photo}
\end{figure}

The air quality monitors used as part of the Kigali deployment, Figure~\ref{fig: ampersand-sensor-photo}, were designed by open-seneca to be low-cost and portable, measuring geotagged $\text{PM}_{2.5}$ data at a cadence of 1~Hz. The air quality monitors are based on an integrated circuit board and embedded software design, based on an ARM Cortex M4 microcontroller (STM32F405RGT6). The integrated circuit board controls an external Sensirion SPS30 particulate matter sensor and includes an on-board GNSS module (u-blox MAX-M8Q), a Sensirion SHTC3 temperature and humidity sensor, a display for data visualisation, and microSD card for data logging. All sensor readings and metadata are written as comma-separated values to an onboard microSD card. Further information on the hardware and software can be found on open-seneca GitHub page\footnote{\url{https://github.com/open-seneca}}.

\begin{figure}
    \centering
    \includegraphics[width=1.0\linewidth]{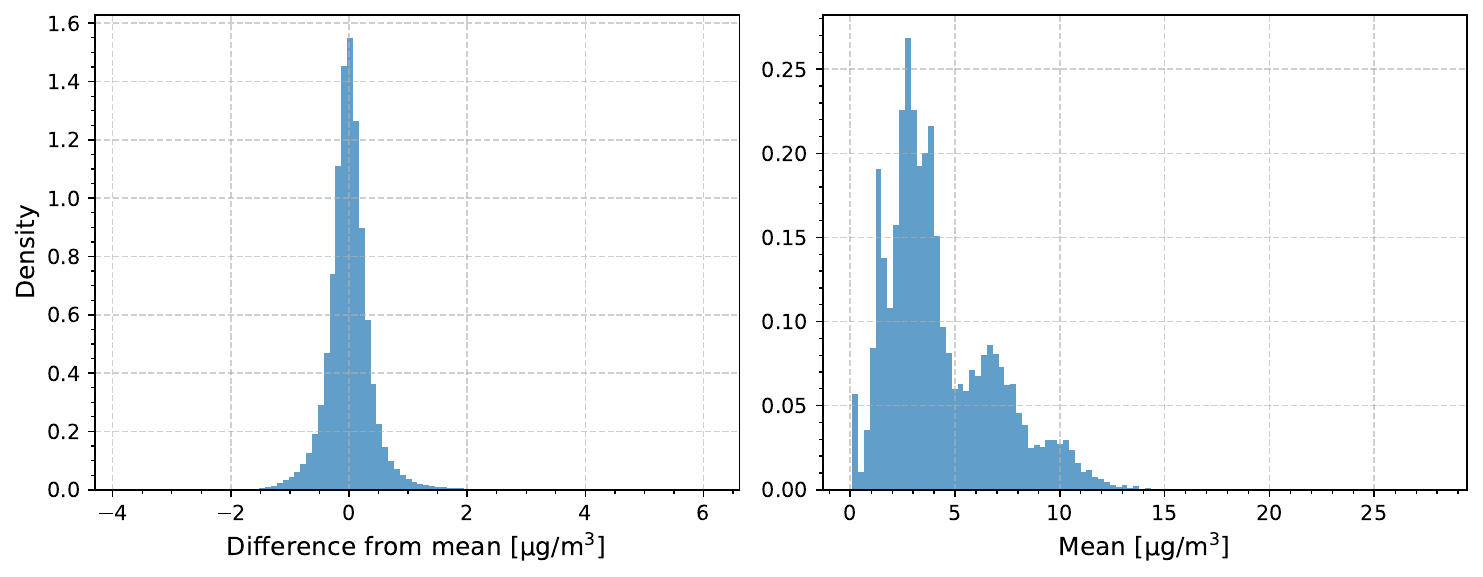}
    \caption{Co-location of 50 sensors over a 2 week period, taking 60~s averages over the whole period, performed in Lisbon, Portugal. Left: Differences of all 50 sensors from the mean over time. Right: Histogram of the mean of all 50 sensors 60~s $\text{PM}_{2.5}$ averages.}
    \label{fig: co-loc-pm25}
\end{figure}

The Sensirion SPS30 particulate matter sensor is factory pre-calibrated.\footnote{\url{https://sensirion.com/products/catalog/SPS30}}. Prior to this study, a co-location of 50 sensors from the same batch as those used in Kigali showed excellent inter-sensor comparability. They achieved $\pm$0.38~\si{\micro\gram/\metre\cubed} standard deviation from the mean over a 2 week period of co-location, as shown in Figure~\ref{fig: co-loc-pm25}, between the range of 0.1 and 28.0~\si{\micro\gram/\metre\cubed}.

Calibration against a reference station was not possible in Kigali due to the absence of a public reference station. However, given the nature of this study and the excellent out-of-the-box inter-sensor comparability, this was not necessary. For reference, Appendix~\ref{appendix: co-location} presents a previous calibration of the same sensor type as those used in Kigali with a reference station at a different location. These calibration data were not used in this study, as the focus was on sensor inter-comparability rather than absolute values.

\subsection{Experiment Setup in Kigali}

The study conducted in Kigali was designed to address air pollution concerns in Rwanda's capital city. It employed 16 open-seneca air quality monitors, mounted on electric delivery motorbikes, part of Ampersand’s Rwanda Ltd fleet. The data collection occurred from 1st of September 2021 to the 1st of October 2021, as part of the Urban Pathways project\footnote{\url{https://web.archive.org/web/20250613013419/http://www.urban-pathways.org/}}, supported by the Wuppertal Institute, UN-Habitat, and the United Nations Environment Programme (UNEP), and funded by the International Climate Protection Initiative (IKI) of the German Ministry of the Environment. 

The study was facilitated through collaboration with the University of Rwanda, the City of Kigali, the Urban Electric Mobility Initiative (UEMI), and Ampersand Rwanda Ltd. The method involved engaging professional delivery drivers as key participants, who carried sensors during their daily delivery routes. This approach ensured efficient data collection and coverage while adhering to the study's time constraints. 

The sensors were mounted at the front of the vehicle, behind the handlebars, and were continuously powered via the electric motorbike’s onboard USB outlet, as shown in Figure~\ref{fig: ampersand-sensor-photo}. In the open-seneca air quality monitor, the inlet of the Sensirion SPS30 $\text{PM}_{2.5}$ sensor faced opposite to the direction of travel to minimise the impact of direct wind on measurements, as the sensors are sensitive to airflow within the sampling channel.

At the conclusion of the data collection period, the microSD cards were manually retrieved from each of the 16 devices. The raw data from each sensor was stored in individual .csv files. These files contained time-series data for PM2.5, temperature, relative humidity, latitude, longitude, and GPS accuracy metrics, recorded at the 1 Hz frequency. The collected raw data was subsequently aggregated, cleaned, and processed to prepare it for spatial and temporal analysis.

\subsection{Data processing} \label{sec: data-processing}

Duplicate rows in the data were removed, as well as any rows where the date, time, latitude or longitude are missing. The data were also filtered to a spatial bounding box encompassing Kigali, with latitude between 2\degree 9\textquotesingle36\textquotedblright S and 1\degree51\textquotesingle00\textquotedblright S, and longitude between 29\degree54\textquotesingle00\textquotedblright E and 30\degree15\textquotesingle00\textquotedblright E. Measurements were filtered to the 1st-20th of September 2021, the most active days of the study, and the temporal granularity of the data was reduced to one-second intervals: for all seconds where a device registered more than one measurement, the first measurement was retained and the rest were discarded. Measurements where speed equals zero were removed, because there were instances where a device remained switched on for several hours while the vehicle was stationary, possibly collecting measurements inside the driver's house rather than the ambient pollution. Observations where $\text{PM}_{2.5}$ $>$ 500\si{\micro\gram\per\cubic\meter} were deemed outliers and removed, with the cut-off value of 500 chosen after consulting typical $\text{PM}_{2.5}$ ranges in polluted cities (e.g., \cite{Sun2022-za}). The processed dataset is publicly available at \citep{https://doi.org/10.5281/zenodo.15206350}.

\begin{figure}
    \centering
    \includegraphics[width=0.9\linewidth]{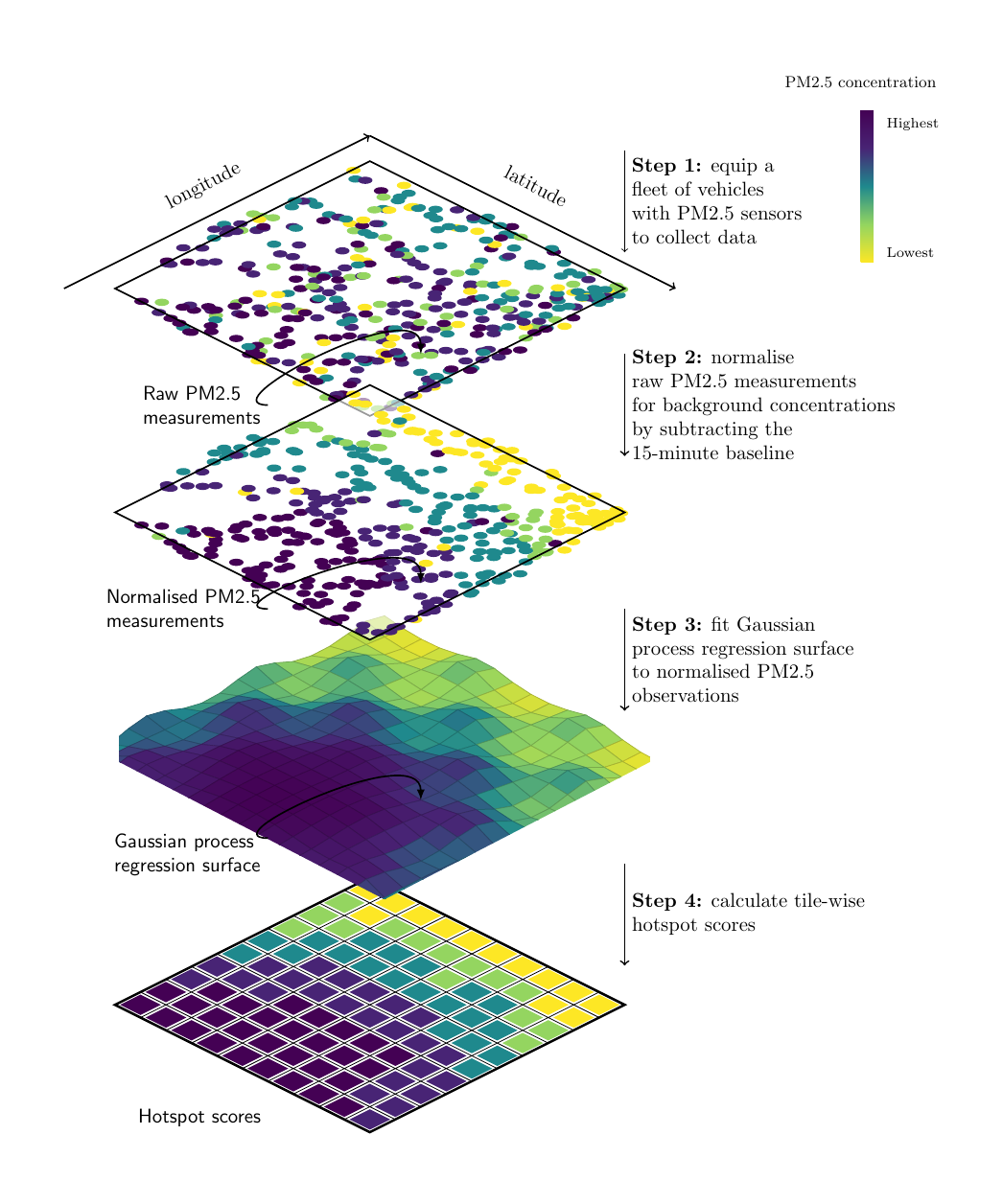}
    \caption{The $\text{PM}_{2.5}$ hotspot identification method proposed in this paper. Raw $\text{PM}_{2.5}$ measurements are collected with low-cost sensors attached to a fleet of vehicles. The measurements are normalised by subtracting the rolling 15-minute median, which removes the hour-to-hour variation in the ambient $\text{PM}_{2.5}$ concentration. A Gaussian process regression surface is fit to the normalised measurements. Finally, tile-wise hotspot scores are calculated to reveal the most polluted locations in the city.}
    \label{fig: pm25-pipeline}
\end{figure}

\subsection{Hotspot detection method} \label{sec: hotspot-detection-method}

Our method for detecting urban $\text{PM}_{2.5}$ hotspots consists of four main steps (Figure \ref{fig: pm25-pipeline}). The first step is to collect $\text{PM}_{2.5}$ concentrations data by equipping a fleet of vehicles with mobile $\text{PM}_{2.5}$ sensors. The collected data should be appropriately processed, e.g., to remove outliers, as in Section \ref{sec: data-processing}. However, we do not prescribe the processing stages because the guidelines should be as general as possible, whereas the data processing required depends on the context of the mobile sensing deployment. After the data is collected and processed, the second step is background normalisation, which entails correcting for the hour-to-hour meteorological variations in ambient $\text{PM}_{2.5}$ concentrations. Mobile monitoring studies typically normalise for background concentrations by scaling the measured values according to the measurements from a local $\text{PM}_{2.5}$ monitoring station \citep{Van_den_Bossche2015-xd, Apte2017-pn, Lim2019-ji}. However, in September 2021, Kigali had no reference $\text{PM}_{2.5}$ monitoring station, and moreover, we want our method to be applicable in other cities with no fixed pollution monitoring apparatus. We therefore propose a background-normalisation procedure that only depends on the mobile sensing data. The raw $\text{PM}_{2.5}$ observations are normalised by selecting a window length $x_w$ and subtracting the median from the previous $x_w$ minutes of observed $\text{PM}_{2.5}$ concentrations (Equation \ref{eq: background-normalisation}). 

\begin{align} \label{eq: background-normalisation}
    y_i &= y_i^{\text{raw}} - \text{median}( \{ y_k^{\text{raw}}: t_i - t_k \in [0, x_w], k \in I\}) 
\end{align}

where

\begin{align*}
y_i^{\text{raw}} &= \text{raw (observed) $\text{PM}_{2.5}$ concentration} \\
    y_i &= \text{normalised $\text{PM}_{2.5}$ concentration} \\
    t_i &= \text{timestamp of $i$'th observation} \\
    I &= \text{index set for mobile sensing observations}
\end{align*}

$x_w$ is an important hyperparameter in our method. A longer window (e.g., $x_w =$ 60 minutes) means more observations are included in the median calculation, so the subtracted baseline is more robust to outliers. Additionally, when the window length is longer the baseline may include measurements from a greater number of drivers, since, at any given second, only some of the drivers are on the road. On the other hand, a shorter window (e.g., $x_w =$ 1 minute) enables normalising for temporal variation at a higher temporal resolution. We compared several different window lengths empirically on the Kigali data ($x_w =$ 1, 5, 15, 30, 60 minutes), and eventually selected $x_w = 15$ minutes because, when grouping observations into 15-minute buckets, the resulting average diurnal profile is smoother than for other choices of $x_w$. Here smoothness is measured as minimising the absolute deviation between measurements in consecutive intervals. In summary, a 15-minute baseline is an appropriate trade-off between robustness to outliers and temporal resolution for the Kigali data.

The third step is to fit a Gaussian process regression surface to the spatial extent of the city. As in the general regression setting, the objective in Gaussian process regression is to learn a function $f$ relating a quantity of interest $y_i$ to a known quantity $\mathbf{x}_i$ in the presence of random noise $\varepsilon_i$:

\begin{equation} \label{eq: gp-regression}
y_i = f(\mathbf{x}_i) + \varepsilon_i
\end{equation}

In our context, $i \in I$ indexes an observation from the data, $\mathbf{x}_i \in \mathcal{X} \subset \mathbb{R}^2$ is the vector of latitude and longitude coordinates, $y_i$ denotes the normalised $\text{PM}_{2.5}$, and $f$ relates the pair of coordinates to its $\text{PM}_{2.5}$ concentration. The Gaussian process model thus only considers the spatial location of an observation, with the background normalisation procedure designed to enable the direct comparison of observations made at different times throughout the month. In Gaussian process regression, the key assumption is that $f$ has a Gaussian process as prior distribution (further technical details are explained in Appendix \ref{appendix: gp}). At any location $\mathbf{x}^*$ in the city extent $\mathcal{X}$, the Gaussian process regression model thus represents the pollution $f(\mathbf{x}^*)$ as a \emph{probability distribution} as opposed to a static value, which we argue is a faithful representation of the real-world dynamics of $\text{PM}_{2.5}$. At any location in an urban environment, the ambient $\text{PM}_{2.5}$ concentration is determined by a multitude of factors such as meteorological conditions, passing vehicles and nearby construction activities, all of which vary in intensity across short time scales. Representing the pollution as a probability distribution rather than a static value quantifies this variability. Additionally, Gaussian process regression outputs spatially-explicit variance estimates as opposed to one single variance estimate for the city, so the model captures the fact that some areas experience a greater variability in pollution levels than others (e.g., a busy intersection versus a rural backstreet).

The fourth and final step is to divide the urban area into a grid of tiles and calculate tile-wise hotspot scores. In line with the probabilistic modelling framework, we employ a frequency-based definition of hotspots. We consider as hotspots the locations where the pollution levels are frequently higher than the average pollution level across the whole city. This definition is compatible with the fact that pollution sources are dynamic, so any location in a city could theoretically experience high pollution at a given time. Therefore, what makes a location a hotspot is that the pollution is regularly greater than the average for the whole city. We propose `hotspot scores' derived from the Gaussian process model to quantify the hotspots. Formally, for a tile with a centroid at $\mathbf{x}_{\text{j}} = ( \text{lat}_{\text{j}}, \text{lon}_{\text{j}})$, we define the hotspot score $h(j)$ as the estimated posterior probability that $f( \mathbf{x}_j)$ exceeds the empirical median of normalised $\text{PM}_{2.5}$ for the city, median($y$):

\begin{equation} \label{eq: hotspot-score}
    h(j) :=  \hat{\mathbb{P}}[ f( \mathbf{x}_j) > \text{median}(y)]
\end{equation}

The hotspot score at $\mathbf{x}_{j}$ quantifies our belief that an observation of the $\text{PM}_{2.5}$ concentration there would exceed the empirical median for the city. For example, a hotspot score of 0.3 means that we expect 30\% of noise-free $\text{PM}_{2.5}$ observations to exceed the city-wide median. All quantities are background-normalised, so the hotspot scores exclude the hour-to-hour variation in ambient $\text{PM}_{2.5}$ concentrations and summarise the spatial variability between locations. A common baseline $\text{median}(y)$ is used for every tile so that the hotspots are relative to the whole city, as opposed to the window-based approaches discussed in the Introduction. This makes the hotspot scores more useful to public health officials, for whom the scope of a clean-air initiative is typically an entire city rather than a small neighbourhood within that city. Appendix \ref{appendix: gp} explains how the hotspot scores are computed in practice.

The tile size regulates the granularity of the air quality estimates. Urban air quality studies with gridded observations typically choose a tile length of tens or hundreds of meters \citep{Wang2023-fa, Apte2017-pn, Hasenfratz2015-yk}. We recommend adjusting the tile size according to the sparsity of the observed data, since sampling density has been shown to affect the accuracy of spatial air pollutant estimates \citep{Van_den_Bossche2015-xd}.

\subsection{Validating the method}

In addition to the $\text{PM}_{2.5}$ concentrations data we gathered in Kigali, we also evaluate our method using another dataset of spatially-explicit average $\text{PM}_{2.5}$ concentrations in Beijing, China. Published by \cite{Wang2023-fa}, this dataset consists of estimated seasonal average $\text{PM}_{2.5}$ concentrations at a $30  \times 30m$ spatial resolution. The authors estimated $\text{PM}_{2.5}$ concentrations by combining the WRF-CMAQ atmospheric dispersion model \citep{Wong2012-qv} with several data sources, including top-of-atmosphere satellite data, land-use data, and terrain data like elevation above sea level. We evaluate our hotspot-detection method by treating their spatial $\text{PM}_{2.5}$ estimates in Beijing as a ground-truth spatial $\text{PM}_{2.5}$ distribution, then simulating a mobile sensing dataset that reflects the spatial average data, as described in Section \ref{sec: data-simulation}, and applying our hotspot-detection method to try and recover the $\text{PM}_{2.5}$ hotspots from the spatiotemporally sparse mobile sensing data. The advantage of evaluating our method on the Beijing data is that we can quantitatively evaluate the hotspot scores with reference to the ground truth, whereas in a real-world data collection, the ground truth is not known.

Two evaluation areas approximately $3 \times 3$km 
in size were selected from the Beijing data. The first evaluation area is in the East of the city and is referred to as `CLT', because the area encompasses a landmark building called the \textbf{C}hina \textbf{L}ife \textbf{T}ower (39\degree55\textquotesingle26\textquotedblright N,  116\degree26\textquotesingle22\textquotedblright E). The second evaluation area is in the Southeast of the city, denoted `LP' because it encompasses the \textbf{L}ongtan \textbf{P}ark (39\degree52\textquotesingle47\textquotedblright N, 116\degree26\textquotesingle25\textquotedblright E). Although \cite{Wang2023-fa} published $\text{PM}_{2.5}$ concentrations estimates for all four seasons, we only used their Autumn estimates.

We selected two different evaluation areas from the Beijing dataset in order to make a more rounded evaluation of our method. Beijing CLT is characterised by spatial homogeneity and a lack of distinct hotspots. The difference between the average $\text{PM}_{2.5}$ concentration of the most and least polluted tiles is only $6.06$\si{\micro\gram\per\cubic\meter} , so we wanted to test how the hotspot scores are affected when the inter-urban variability is low. By contrast, Beijing LP has one very distinctive hotspot along the bottom and right edges of the quadrant, and another hotspot with an unusual figure-of-eight shape in the Northwest. There are also sharp boundaries between areas with higher and lower pollution throughout the quadrant. In Beijing LP, we were interested in whether the hotspot-detection method can recreate the unusual geometry of the pollution hotspots and identify the sharp boundaries between neighbourhoods.

\subsubsection{Simulating the mobile sensing data} \label{sec: data-simulation}

Recall that step one of our method is data collection with mobile sensors. However, for this evaluation, instead of deploying mobile sensors in the city, we instead simulated the mobile sensing data collection stage. This allows us to ignore any errors between the $\text{PM}_{2.5}$ estimates and the real $\text{PM}_{2.5}$ concentrations in Beijing. Our objective is not to model $\text{PM}_{2.5}$ concentrations faithfully in Beijing, but rather to evaluate whether our method can recover the hotspots in the data.

Simulating the mobile sensing $\text{PM}_{2.5}$ measurements began with simulating the vehicle routes. Using road network data from the Open Street Map (OSM), we selected ten thousand pairs of random on-road points within the evaluation area. Every pair of points represents a start and end location of a driver's journey: $\{ ((\text{lat}_{s}^{\text{start}}, \text{lon}_{s}^{\text{start}}), (\text{lat}_{s}^{\text{end}}, \text{lon}_{s}^{\text{end}})) \text{ for } s = 1, 2, \ldots, 10,000 \}$. The \verb|networkx| package in Python was then used to compute the shortest on-road path between every pair of points \citep{Hagberg2008-yo}. This produces ten thousand on-road routes, which roughly corresponds to sixteen delivery drivers making thirty trips per day for twenty days ($20 \times 16 \times 30 = 9600$). 

The next step was to generate observation locations and associated timestamps along all the routes. Recall that the mobile sensors measure the $\text{PM}_{2.5}$ concentration once per second. Assuming a constant speed of 30km/h, this corresponds to one observation every 8.33 meters, so observation locations were generated at uniform intervals of 8.33 meters along the on-road routes. Every start location was assigned a random timestamp in the period of September 2019, then timestamps were assigned in increments of one second to the subsequent observations within the same journey. This resulted in a total of 2.7 million observation locations of the form $(\text{lat}, \text{lon}, t)$ for each evaluation area, equivalent to approximately $750$ hours of data. 

Finally, a $\text{PM}_{2.5}$ measurement was assigned to every $(\text{lat}, \text{lon}, t)$ tuple according to the following formula:

\begin{equation} \label{eq: synthetic-pm25}
    y_i^{raw} = hm_{t_i} \times \text{PM}_{2.5}^{j} + \kappa_i \gamma_i + \varepsilon_i
\end{equation}
where
\begin{align*}
    j &= \text{index of tile containing observation } i\\ \text{PM}_{2.5}^{j} &= \text{average $\text{PM}_{2.5}$ concentration in tile j} \\
    hm_{t_i} &= \text{hourly multiplier for time } t_i \\
    \varepsilon_i, \kappa_i, \text{ and } \gamma_i&= \text{stochastic noise variables}\\ 
\end{align*}

\begin{figure}
    \centering
    \includegraphics[width=0.5\linewidth]{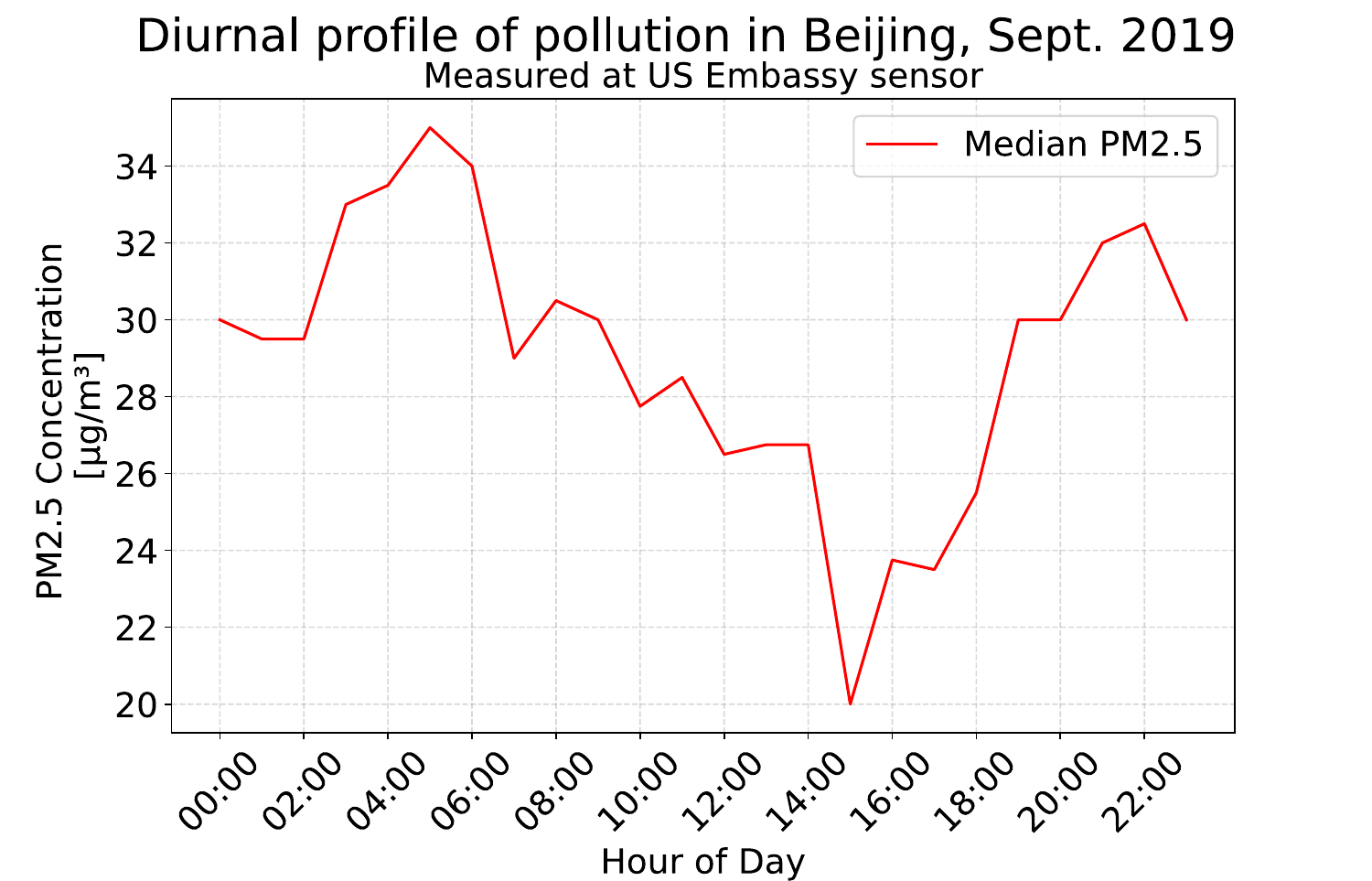}
    \caption{Median diurnal profile of $\text{PM}_{2.5}$ concentrations measured by the US embassy in Beijing (September 2019).}
    \label{fig: beijing-diurnal-profile}
\end{figure}

The hourly multiplier term $hm_{t_i}$ captures the fact that depending on the time of day, a driver passing through the same location can experience drastically different pollution levels due to traffic patterns or meteorological variations. Hourly multipliers were computed using data from the US embassy's stationary $\text{PM}_{2.5}$ sensor in Beijing (Figure \ref{fig: beijing-diurnal-profile}). The sensor measures the ambient $\text{PM}_{2.5}$ once every hour, and we converted the embassy measurements to hourly multipliers by dividing each measurement by the overall median of the embassy measurements for September 2019. As an example, an hourly multiplier of 1.2 means that, in a given hour on a given day, the $\text{PM}_{2.5}$ concentration was 20\% higher than the overall median for September 2019. By scaling according to the embassy measurements, we introduce realistic temporal variation into the synthetic observations.

Remaining variables in Equation \ref{eq: synthetic-pm25} are stochastic: $\varepsilon_i \sim N(0,1)$, $\kappa_i \sim \text{Bern}(0.05)$, and $\gamma_i \sim \text{Gamma}(5,5)$. $\kappa_i \gamma_i$ is a stochastic term that only affects 5\% of the measurements, and corrupts them with an additive $\text{PM}_{2.5}$ value drawn from the $\text{Gamma}(5,5)$ distribution. The $\text{Gamma}(5,5)$ distribution has a positive support and a long tail, so this term represents sudden pollution spikes like the exhaust emissions from a passing truck or dust from unpaved roads, a phenomenon observed in our Kigali data and also reported in previous mobile sensing campaigns \citep{Van_den_Bossche2015-xd}. $\varepsilon$ is always present and represents random noise, including possible measurement error. These stochastic interferences further complicate the recovery of the spatial $\text{PM}_{2.5}$ hotspots.

\subsubsection{Evaluation plots and metrics} \label{sec: eval-plots-and-metrics}

We address two main questions when evaluating our method on the simulated Beijing data:

\begin{itemize}
    \item Do the hotspot scores reflect the spatial profile of average $\text{PM}_{2.5}$ concentrations throughout the city?
    \item{Are the hotspot scores probabilistically well-calibrated?}
\end{itemize}

To address the first question, we plot the spatial distributions of the hotspot scores and of the gridded average $\text{PM}_{2.5}$ concentrations side by side. This helps visualize whether the method can successfully recreate the spatial pollution profile in a city from spatiotemporally sparse mobile sensing data. We also compute the Spearman rank correlation coefficient $\rho(h, \text{PM}_{2.5})$ between the hotspot scores $h := (h(j))_{j \in J}$ and the spatial average pollution values, $\text{PM}_{2.5} := (\text{PM}^j_{2.5})_{j \in J} $:

\begin{align*}
    \rho (h, \text{PM}_{2.5} ) &:= \frac{\text{Cov}(R(h), R(\text{PM}_{2.5}))}{\sigma_{R(h)} \cdot \sigma_{R(\text{PM}_{2.5})}}
\end{align*}

Here $R(h)$ and $R(\text{PM}_{2.5} )$ are the rank vectors of the hotspot scores and the spatial average $\text{PM}_{2.5}$ values, respectively, while the denominator terms are the standard deviations of these rank vectors. Spearman's rank correlation coefficient $\rho$ is a non-parametric measure of correlation which is robust to non-linearity. It is bounded between 1, which means a perfect positive correlation, and -1, which means a perfect negative correlation. The closer $\rho$ is to 1, the better the hotspot scores are at capturing the spatial pollution profile.

To address the second question: recall that the hotspot scores are estimated posterior probabilities for the event $\{f(\mathbf{x}_i) > \text{median(y)}\} $ (Equation \ref{eq: hotspot-score}). In the simulated data, $f(\mathbf{x}_i)$ is known for every observation $\mathbf{x}_i$, as determined by the simulation process (Equation \ref{eq: synthetic-pm25}). Hence we can perform a calibration analysis, comparing the occurrence of the event $\{f(\mathbf{x}_i) > \text{median(y)}\} $ against the hotspot scores, which are the spatially-explicit estimated probabilities of this event. Perfect calibration would mean that in every tile, the hotspot score equals the proportion of observations where the event $\{f(\mathbf{x}_i) > \text{median(y)}\} $ occurs, which we henceforth call the exceedance proportion. 

To perform a calibration analysis, we split the mobile sensing data into a test set, containing observations from six randomly-selected days in September 2019, and a training set, containing the observations from the remaining twenty four days. Hotspot scores are computed using only the training data, then the calibrations are evaluated on the test data. Test observations are grouped into ten uniformly-spaced bins according to the hotspot score of their tile, i.e.,

\begin{align*}
    \text{bin}_1 &= \{ (\mathbf{x}_i, y_i) \in \mathcal{X} \times \mathcal{Y}: h(\mathbf{x}_i) \in [0, 0.1] \} \\
    \text{bin}_2 &= \{ (\mathbf{x}_i, y_i) \in \mathcal{X} \times \mathcal{Y}: h(\mathbf{x}_i) \in (0.1, 0.2] \} \\
        \vdots\\
    \text{bin}_{10} &= \{ (\mathbf{x}_i, y_i) \in \mathcal{X} \times \mathcal{Y}: h(\mathbf{x}_i) \in (0.9, 1.0] \} \\
\end{align*}

which enables plotting reliability diagrams to show the actual exceedance proportion among observations in each bin. We also report the Brier score:

\begin{align*}
    \text{Brier score} &= \frac{1}{N} \sum_{i=1}^{N}(h(\mathbf{x}_i) - b_i)^2 \\
    \text{where } b_i &= \begin{cases}
    1 & f(\mathbf{x}_i) > \text{median}(y) \\
    0 & f(\mathbf{x}_i) \leq \text{median}(y) 
    \end{cases} \\
    \text{and } h(\mathbf{x}_i) &= \text{hotspot score of tile containing } \mathbf{x}_i
\end{align*}

which is the average squared error between the hotspot scores and the binary variable $b_i$ for whether the $i$'th observation exceeds the median. Additionally, we report the expected calibration error (ECE):

\begin{align*}
    \text{ECE} &= \sum_{k=1}^{10} \frac{|\text{bin}_k|}{10} |\text{exc}(\text{bin}_k) - \text{conf}(\text{bin}_k)|  \\
    \text{where } \text{exc}(\text{bin}_k) &= \frac{|\{ (\mathbf{x}_i, y_i) \in \text{bin}_k: f(\mathbf{x}_j) > \text{median}(y) \}|}{|\text{bin}_k|} \\
    \text{and } \text{conf}(\text{bin}_k) &= \frac{1}{|\text{bin}_k|} \sum_{i : (\mathbf{x}_i, y_i) \in \text{bin}_k} h(\mathbf{x}_i)
\end{align*}

which is the mean absolute difference between the exceedance proportion and the average hotspot score across the ten bins \citep{PakdamanNaeini2015}. 

\subsubsection{Calibration regression}

As part of the analysis we also evaluate whether a calibration regression can improve the probabilistic calibration. A calibration regression is a model regressing the observed frequencies in the data on the predicted probabilities. It is commonly applied to transform the outputs of a model that predicts probabilities in the supervised learning setting, with the aim to improve the accuracy of the predicted probabilities \citep{NiculescuMizil2005, Zadrozny2002}.

As argued by \cite{Dormann2020}, a calibration regression is not necessary when the goal of the hotspot-detection is merely to \emph{rank} the tiles by their average pollution. Calibration regression procedures are generally monotone transformations, so rankings are not affected except where ties are introduced. However, a calibration regression is important when one wants to interpret the hotspot scores as posterior probabilities. In Section \ref{sec: results} we investigate whether an out-of-the-box isotonic regression transformation improves the calibration \citep{Barlow1972, NiculescuMizil2005}. Isotonic regression entails fitting an isotonic (monotonic-increasing) function $m$ to transform the hotspot scores $h_i$ into calibrated hotspot scores $h_i^{\text{cal}}$:

\begin{equation*}
    h_i^{\text{cal}} = m(h_i) + \varepsilon_i
\end{equation*}

Using the \verb|sklearn| implementation of the isotonic regressor \citep{scikit-learn}, we investigate whether this transformation reduces the expected calibration error of the hotspot scores in the simulated data, and discuss how such a procedure could be replicated in future mobile monitoring studies.

\FloatBarrier

\section{Results} \label{sec: results}

\subsection{Kigali Data}

\begin{figure}[h!]
    \centering
    \includegraphics[width=0.6\textwidth]{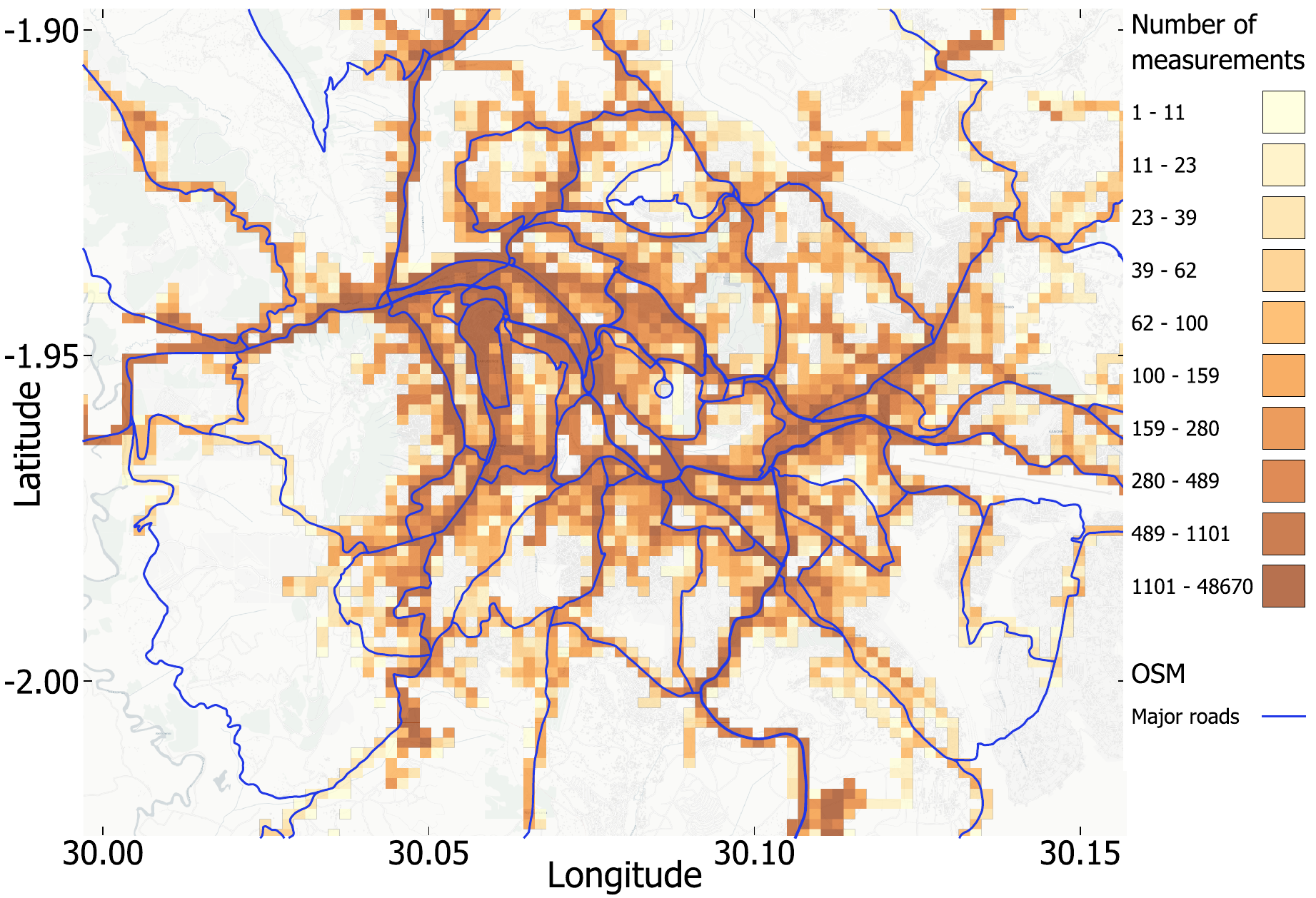}
    \caption{Geographic distribution of $\text{PM}_{2.5}$ measurements in Kigali. Measurements are concentrated in the city centre and along major roads.}
    \label{fig: kigali-n-measurements}
\end{figure}

After processing, the Kigali data consists of 2.02 million observations of the $\text{PM}_{2.5}$ concentration in and around Kigali. With one measurement per second, this corresponds to $>562$ hours of data. Figure \ref{fig: kigali-n-measurements} illustrates the geographic distribution of the observed measurements. In general, grid squares with the greatest number of measurements encompass a stretch of a major road, reflecting that major roads are often convenient for a driver taking the fastest route from A to B. Additionally, the measurements are concentrated in the city centre more than the outskirts. 

\begin{figure}[h!]
    \centering

    \begin{subfigure}[b]{0.45\textwidth}
        \centering
        \includegraphics[width=\textwidth]{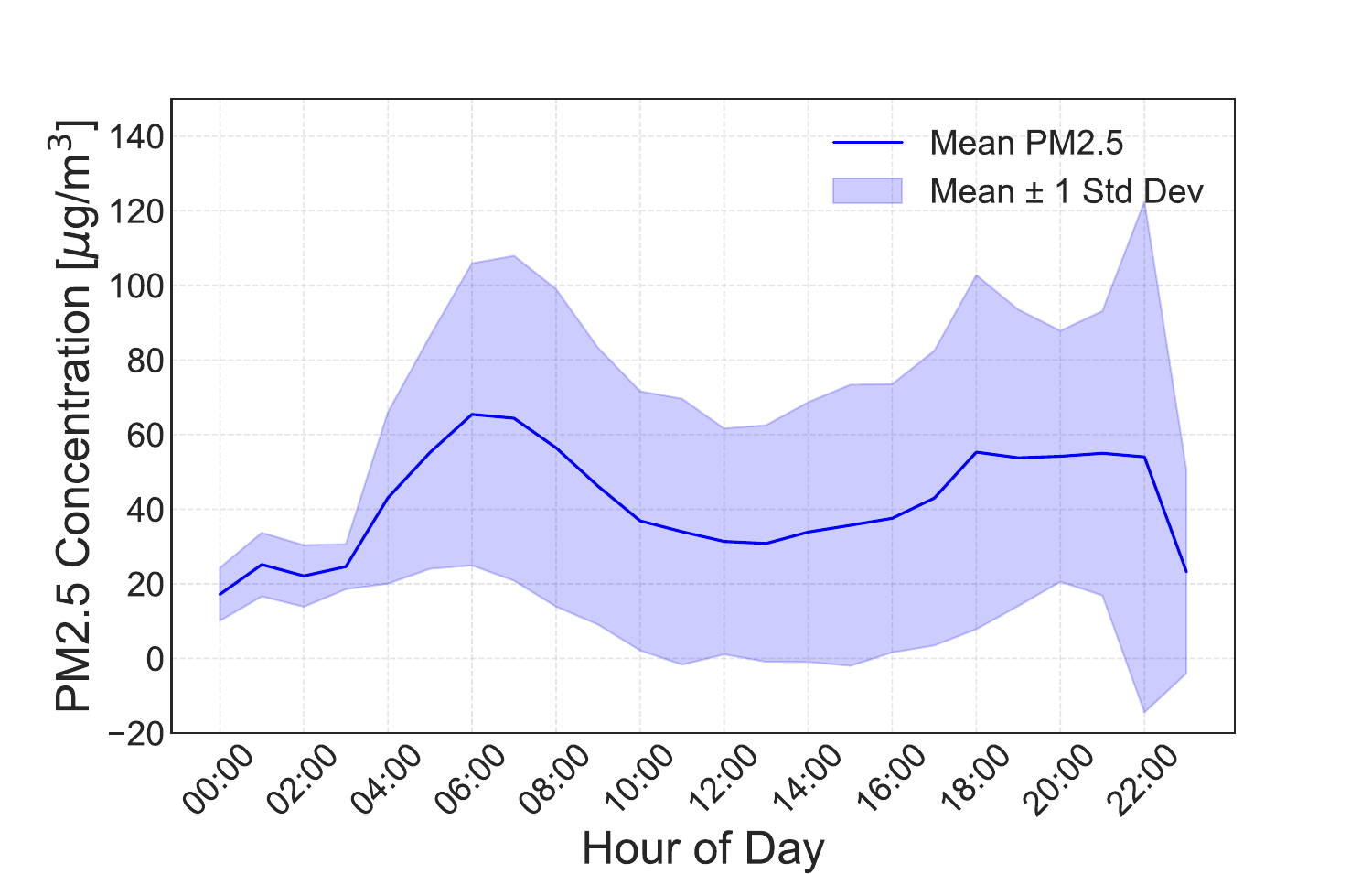}
        \caption{Average diurnal profile.}
        \label{fig: kigali-diurnal-pm25}
    \end{subfigure}
    \hfill
    \begin{subfigure}[b]{0.45\textwidth}
        \centering
        \includegraphics[width=\textwidth]{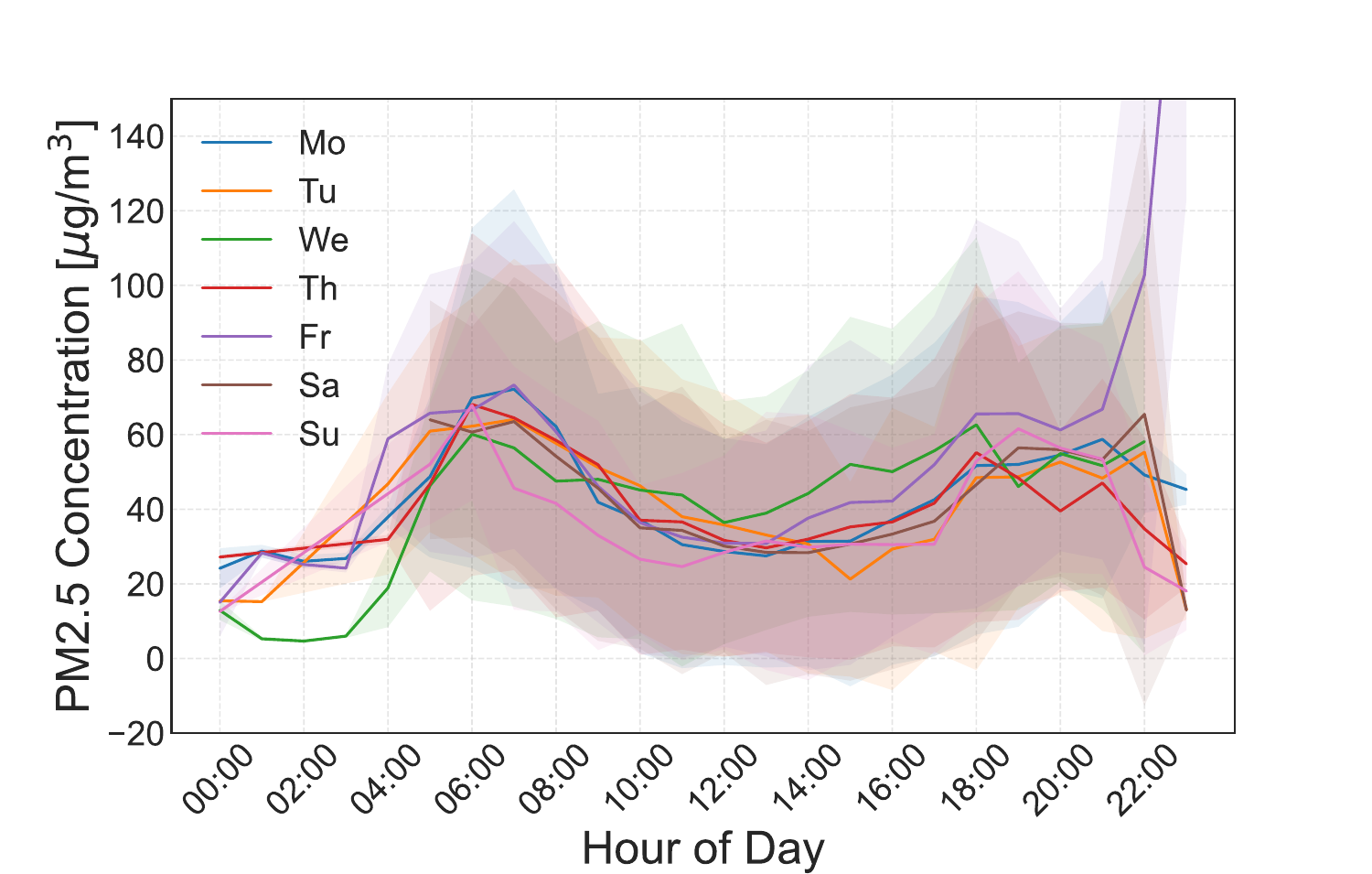}
        \caption{Average diurnal profile, separated by weekday.}
        \label{fig: kigali-diurnal-pm25-dotw}
    \end{subfigure}

    \caption{Diurnal $\text{PM}_{2.5}$ profile in Kigali: (a) overall average and (b) averages by day of the week. The daily pattern of PM\textsubscript{2.5} concentrations was observed to be fairly similar across all days of the week. Times are given in the local timezone (Central African Time, UTC+02:00).}
    \label{fig: kigali-diurnal-comparison}
\end{figure}

Figure \ref{fig: kigali-diurnal-comparison} shows the temporal pattern of $\text{PM}_{2.5}$ concentrations in our dataset. The average $\text{PM}_{2.5}$ concentration exhibits a diurnal pattern, with the mean concentration peaking in the early morning and again in the late afternoon and evening (Figure \ref{fig: kigali-diurnal-pm25}). The standard deviation of the $\text{PM}_{2.5}$ measurements is smallest at night (00:00 - 04:00) and larger throughout the rest of the day. This diurnal pattern is similar on every day of the week, and closely matches the pattern observed in previous $\text{PM}_{2.5}$ measurement studies in Kigali \citep{subramanian2020air, Kagabo2018-sx}. 

An interesting feature is that during the morning (07:00-11:00), average $\text{PM}_{2.5}$ is clearly lower on Sundays than on any other weekday, but from 12:00 onwards the $\text{PM}_{2.5}$ level on Sunday resembles the other weekdays again. This coincides with the timing of the `car-free Sundays' scheme in Kigali. On the first and third Sunday of every month, several main roads in Kigali are closed to motorized vehicles between 07:00 - 10:00, and pedestrians are encouraged to gather in the vacant streets for group exercise sessions. At 10:00 all roads are re-opened. Two out of the three Sundays within our data collection campaign were `car-free', which could plausibly affect the air quality encountered by the mobile sensors, although more robust evidence would be needed to establish a causal link between the scheme and decreased ambient pollution levels. Figure \ref{fig: kigali-diurnal-pm25-dotw} also suggests that the mean $\text{PM}_{2.5}$ concentration increases sharply on Friday nights and dips during the early hours of Wednesday morning. However, these extremes both correspond to a small number of measurements, so probably do not reflect a typical trend.

We argue that because our data is gathered while the drivers carry out their usual jobs - as opposed to driving along pre-determined routes - the average $\text{PM}_{2.5}$ concentrations recorded are representative of the $\text{PM}_{2.5}$ exposure of Kigali's road users. It is therefore alarming to see that the mean $\text{PM}_{2.5}$ concentration exceeds the WHO's recommended level of $5$\si{\micro\gram\per\cubic\meter} at all hours of the day. The overall mean $\text{PM}_{2.5}$ concentration in the Kigali data is $44.61$\si{\micro\gram\per\cubic\meter} with a standard deviation of $39.47$\si{\micro\gram\per\cubic\meter}. Furthermore, 91.9\% of $\text{PM}_{2.5}$ concentration observations exceed $15$\si{\micro\gram\per\cubic\meter}, whereas the WHO guidelines recommend that $\text{PM}_{2.5}$ exposure should not exceed $15$\si{\micro\gram\per\cubic\meter} for any more than a handful of days in the year. Our data supports previous studies finding that air pollution in Kigali often exceeds the safe limits set by the WHO \citep{subramanian2020air, Kagabo2018-sx, Henninger2013-is}, and provides evidence that the air quality encountered by road users is dangerously unhealthy as a result. 

\subsection{Hotspot Detection in Kigali}
\begin{figure}[h!]
    \centering

    \begin{subfigure}[t]{0.45\textwidth}
        \centering
        \includegraphics[width=\textwidth]{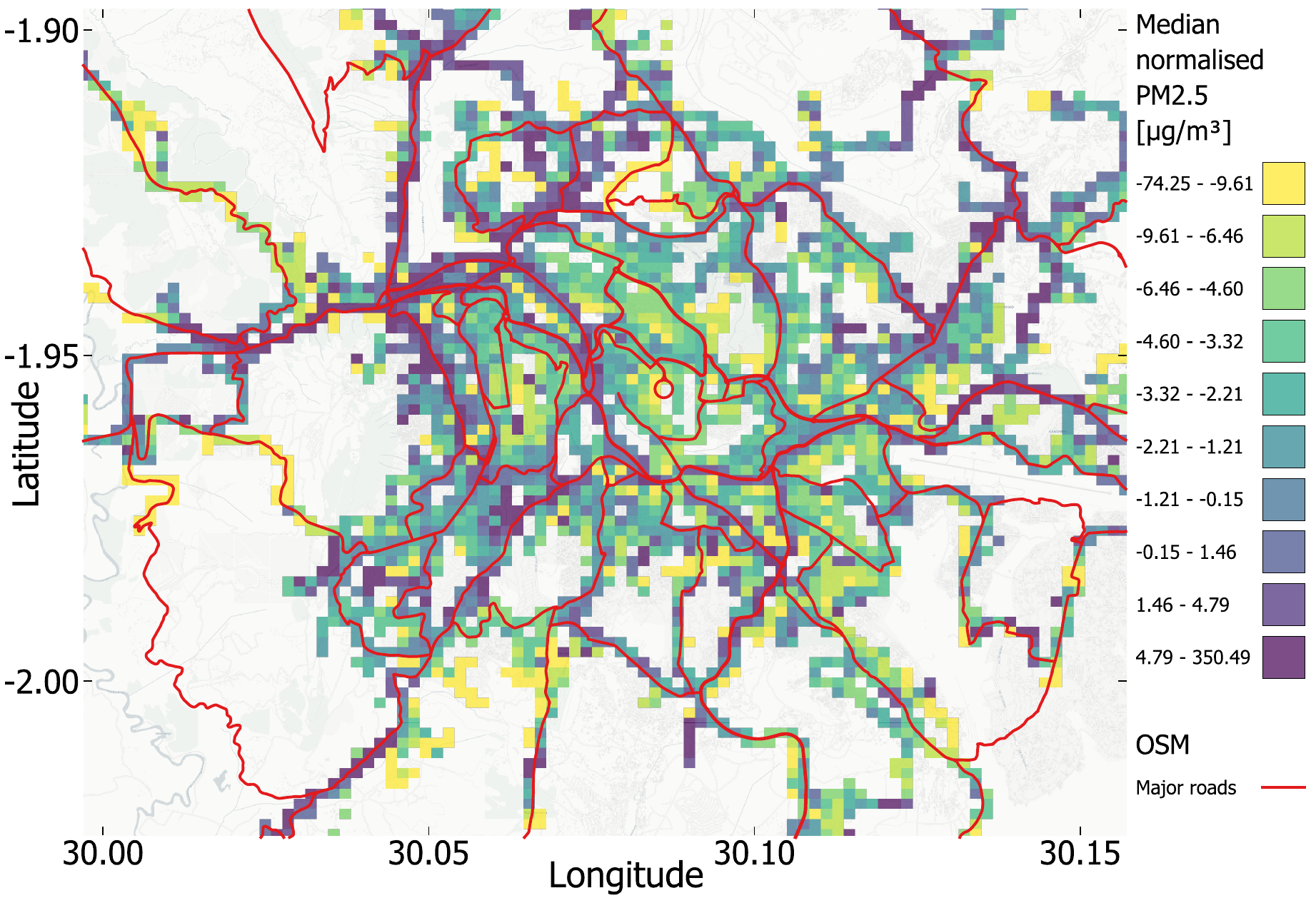}
        \caption{Median normalised $\text{PM}_{2.5}$ concentration per tile.}
        \label{fig:kigali-pm25-distribution}
    \end{subfigure}
    \hfill
    \begin{subfigure}[t]{0.45\textwidth}
        \centering
        \includegraphics[width=\textwidth]{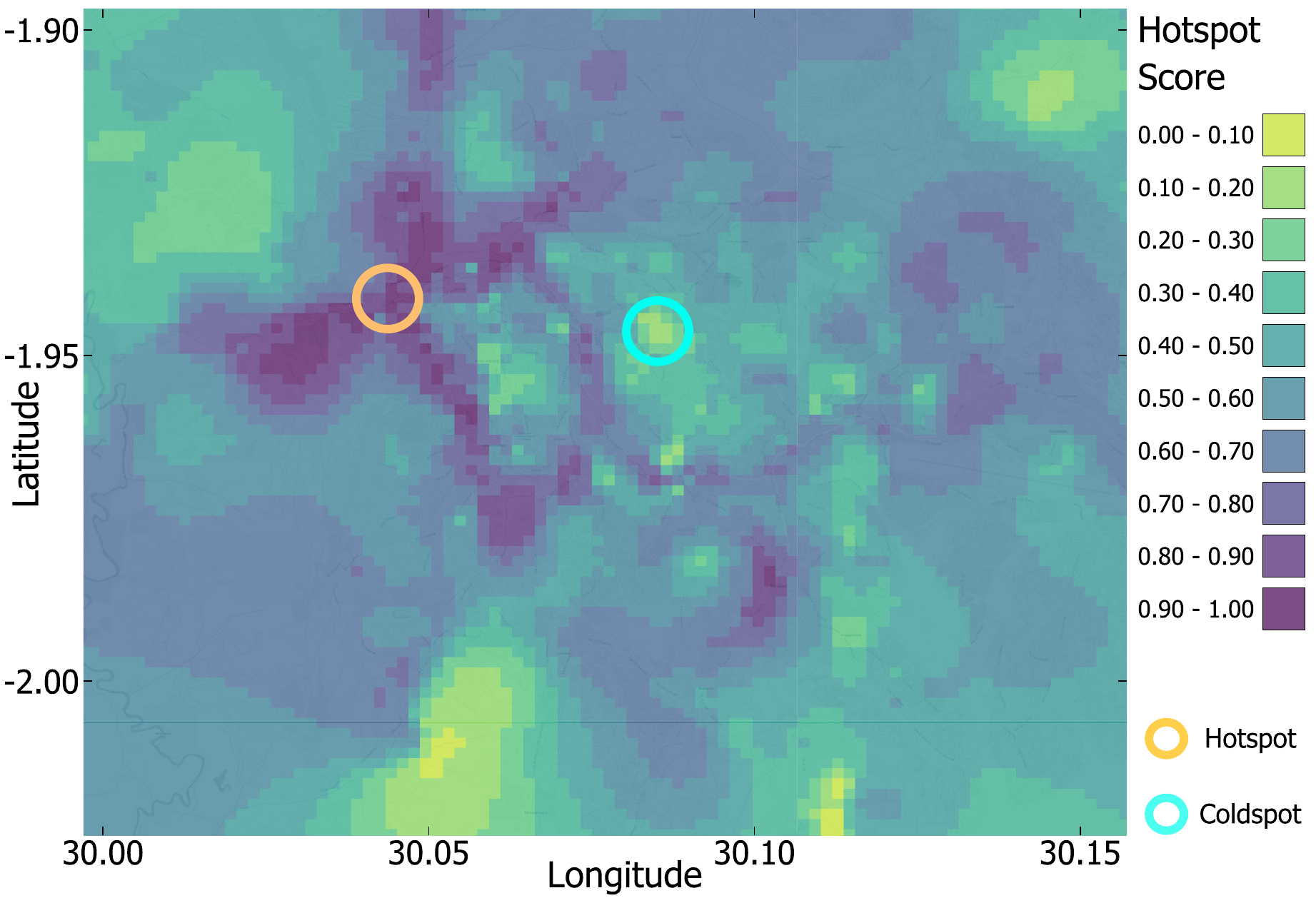}
        \caption{Hotspot scores.}
        \label{fig:kigali-hotspot-scores}
    \end{subfigure}
    
    \vspace{0.5cm} 

    \begin{subfigure}[t]{0.45\textwidth}
        \centering
        \includegraphics[width=\textwidth]{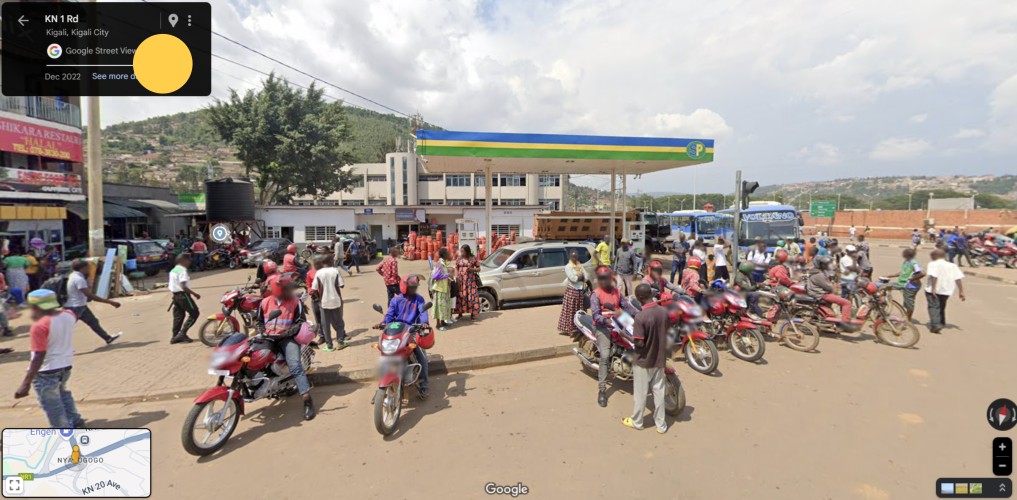}
        \caption{1\degree56\textquotesingle31\textquotedblright S, 30\degree2\textquotesingle38\textquotedblright E (December 2022).}
        \label{fig:kigali-hotspot-google-image}
    \end{subfigure}
    \hfill
    \begin{subfigure}[t]{0.45\textwidth}
        \centering
        \includegraphics[width=\textwidth]{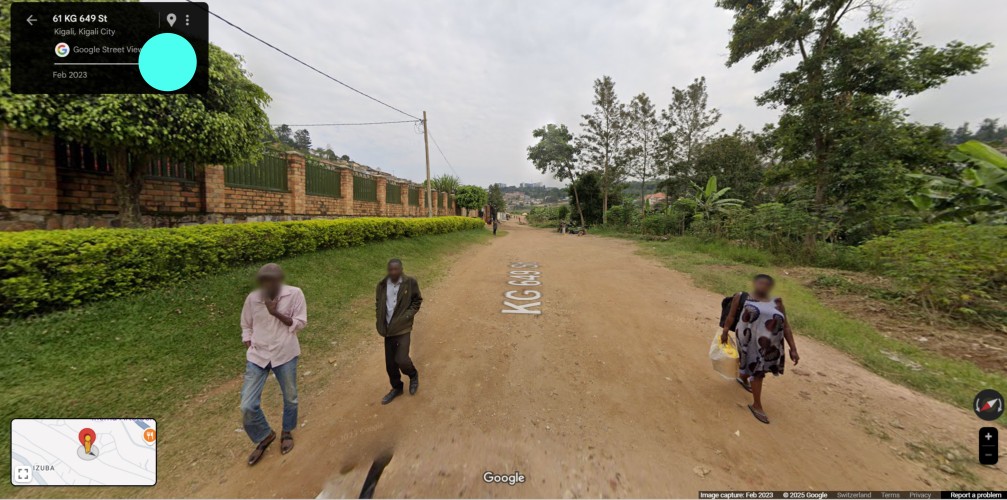}
        \caption{1\degree56\textquotesingle45.7\textquotedblright S, 30\degree05\textquotesingle05.9\textquotedblright E (February 2023).}
        \label{fig:kigali-coldspot}
    \end{subfigure}

    \caption{Normalised $\text{PM}_{2.5}$ concentrations (a), hotspot scores (b), and illustrative Google Street View images from a hotspot (c) and coldspot (d).}
    \label{fig:kigali-pm25-analysis}
\end{figure}

We proceeded to apply our hotspot-detection method to the mobile sensing data in order to estimate the $\text{PM}_{2.5}$ hotspot locations in Kigali. Previous air quality studies in Kigali have relied on ground measurements at fixed monitoring stations (Subramanian et al., 2020; Kagabo, 2018). One study also measured particulate matter concentrations using mobile sensors, but they focused on PM10 and followed a prescribed route (Henniger, 2013). Therefore, to the best of our knowledge, ours is the first study to present the spatial distribution of $\text{PM}_{2.5}$ pollution throughout the entire city of Kigali. As a spatial resolution we chose $200 \times 200$m as a trade-off between a high spatial granularity and ensuring that the tiles are represented by a reasonable number of observations. 

Figure \ref{fig:kigali-pm25-distribution} displays the median normalised $\text{PM}_{2.5}$ value per tile together with the network of major roads in Kigali retrieved from OSM. In general, the tiles with the highest normalised $\text{PM}_{2.5}$ values encompass a stretch of a major road, suggesting that road traffic is an important component of the spatial variability of air pollution within Kigali. This is consistent with the findings of \cite{subramanian2020air}, who concluded that background atmospheric $\text{PM}_{2.5}$ levels in Kigali are mostly driven by biomass burning practices and seasonal weather patterns, whereas day-to-day pollution is mostly driven by traffic. 

The hotspot scores outputted by the Gaussian process regression model are variegated (Figure \ref{fig:kigali-hotspot-scores}), reflecting the flexibility of Gaussian process regression models. The hotspot scores suggest that a large hotspot in Kigali is centered around 1\degree56\textquotesingle31\textquotedblright S,  30\degree2\textquotesingle36\textquotedblright E, which is the intersection between four major roads: the KN1, KN7, KN20 and Kigali-Gatuna. Here, the hotspot scores are among the largest for the whole city, with hotspot scores $>0.9$ in several neighbouring tiles, indicating that the model predicts that ambient $\text{PM}_{2.5}$ concentrations in this area often exceed the city-wide median. Google Street View images of this neighbourhood suggest that this is a sensible finding  (Figure \ref{fig:kigali-hotspot-google-image}). One sees multiple lanes of traffic, several petrol stations in the neighbourhood, and pockets of exposed sand (e.g., at the intersection between the KN1 and Kigali-Gatuna roads), with the latter a known contributor of non-exhaust traffic emissions \citep{Askariyeh2020-iq}. High hotspot scores are continually observed along the four major roads for at least a kilometer as they depart from this intersection. A second prominent hotspot is in the South of the Kicukiro neighbourhood, around 1\degree59\textquotesingle 01.9 \textquotedblright S 30\degree06\textquotesingle08.7\textquotedblright E. Two neighbouring tiles have hotspot scores $>0.9$, encompassing both a stretch of the multi-lane KK15 highway and a local tyre retreading company. While the purpose of our study is not to identify pollution \textit{sources}, it is nevertheless encouraging that the locations of major hotspots identified in Kigali coincide with known sources of $\text{PM}_{2.5}$ pollution. For comparison, Google Street View images were also inspected at a less polluted area along the KG 649 street (Figure \ref{fig:kigali-coldspot}, around 1\degree56\textquotesingle45.7\textquotedblright S 30\degree05\textquotesingle05.9\textquotedblright E). The model assigned low hotspot scores to this street (0.1-0.3). Google Street View images suggest that this is a low-traffic area, with few visible vehicles and open fields directly bordering the road, rather than denser developments like residential housing. Furthermore, the road is lined with trees, which are associated with cleaner air through dry deposition processes \citep{Hirabayashi2016-hy}. Low hotspot scores seem reasonable for this area.

\subsection{Hotspot detection in Beijing}

\begin{figure}
    \centering
    \begin{tabular}{m{4cm} m{5cm} m{4cm}}
        \centering \textbf{\ \ Average $\text{PM}_{2.5}$ \newline Concentration} & \centering \textit{Mobile Sensing \& Hotspot prediction} & \centering \textbf{Hotspot\newline scores} \\
    \end{tabular}

    \begin{tabular}{m{6cm} m{1cm} m{6cm}}
        \centering
        \begin{subfigure}{6cm}
            \includegraphics[width=\linewidth]{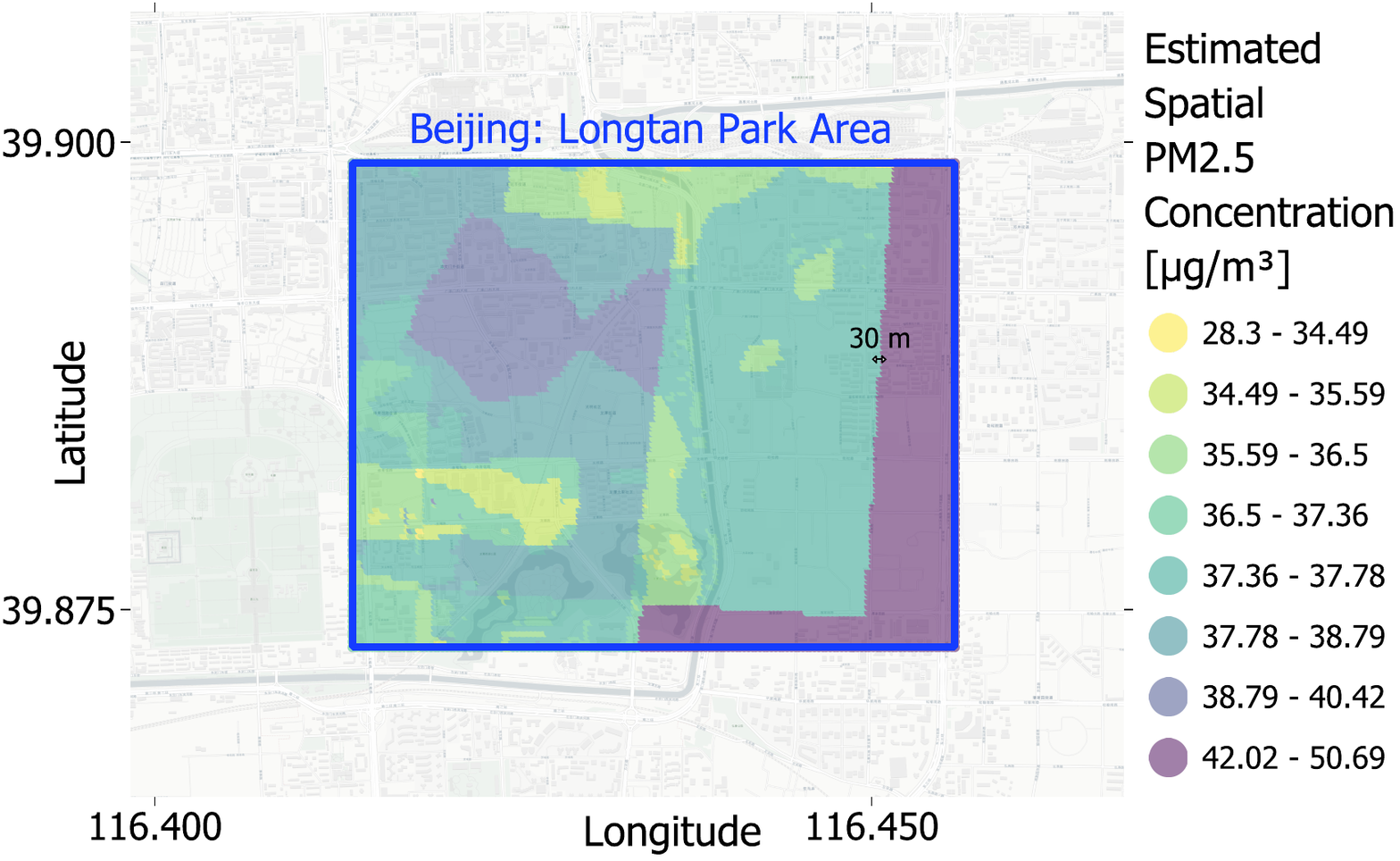}
            \caption{}
            \label{fig: beijing-lp-spatial}
        \end{subfigure}
        &
        \centering
        \begin{tikzpicture}
            \draw[->, thick] (0,0) -- (1,0);
        \end{tikzpicture}
        &
        \centering
        \begin{subfigure}{6cm}
            \includegraphics[width=\linewidth]{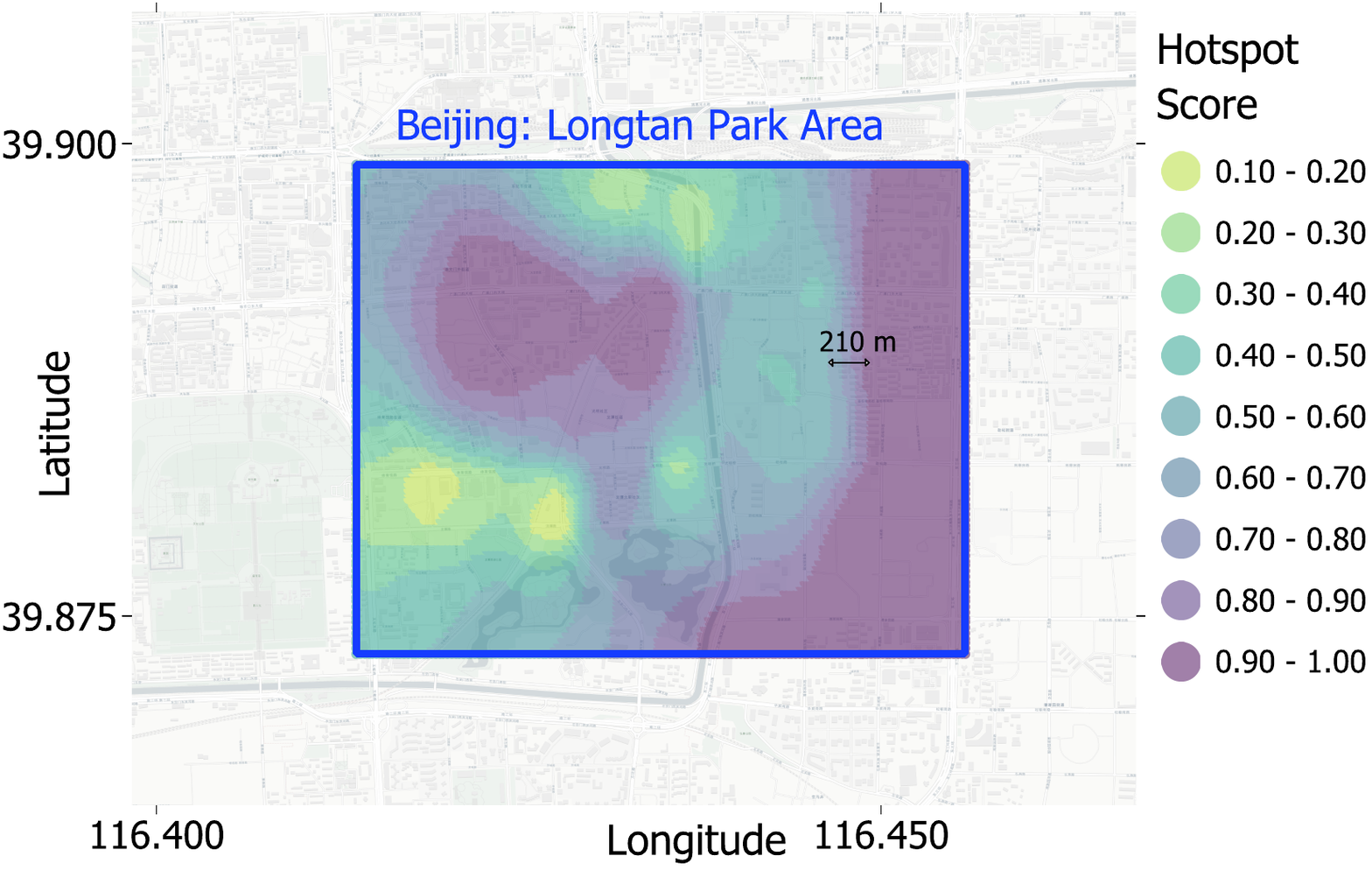}
            \caption{}
            \label{fig: beijing-lp-hotspot-scores}
        \end{subfigure}
    \end{tabular}

    \begin{tabular}{m{6cm} m{1cm} m{6cm}}
        \centering
        \begin{subfigure}{6cm}
            \includegraphics[width=\linewidth]{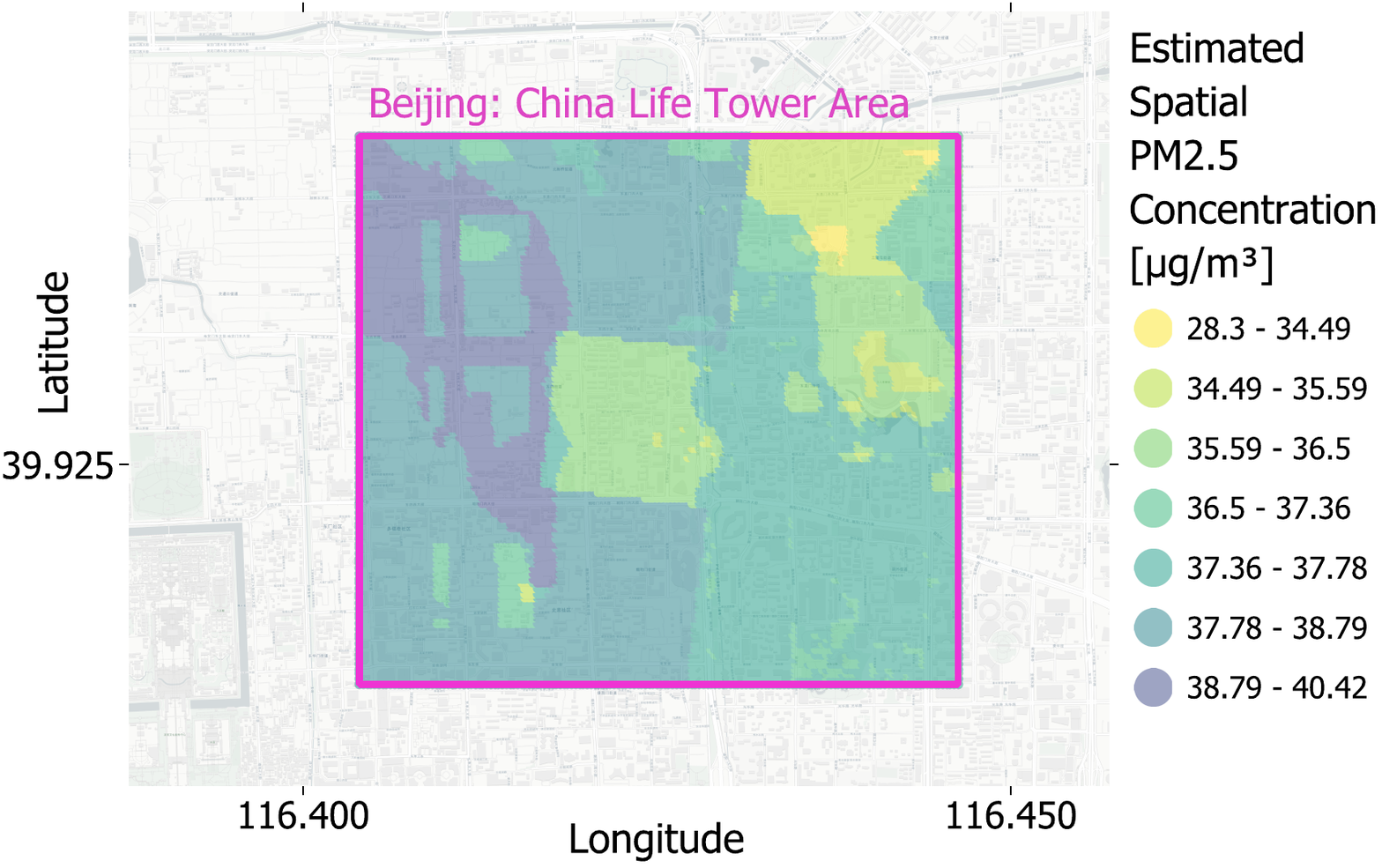}
            \caption{}
            \label{fig: beijing-clt-spatial}
        \end{subfigure}
        &
        \centering
        \begin{tikzpicture}
            \draw[->, thick] (0,0) -- (1,0);
        \end{tikzpicture}
        &
        \centering
        \begin{subfigure}{6cm}
            \includegraphics[width=\linewidth]{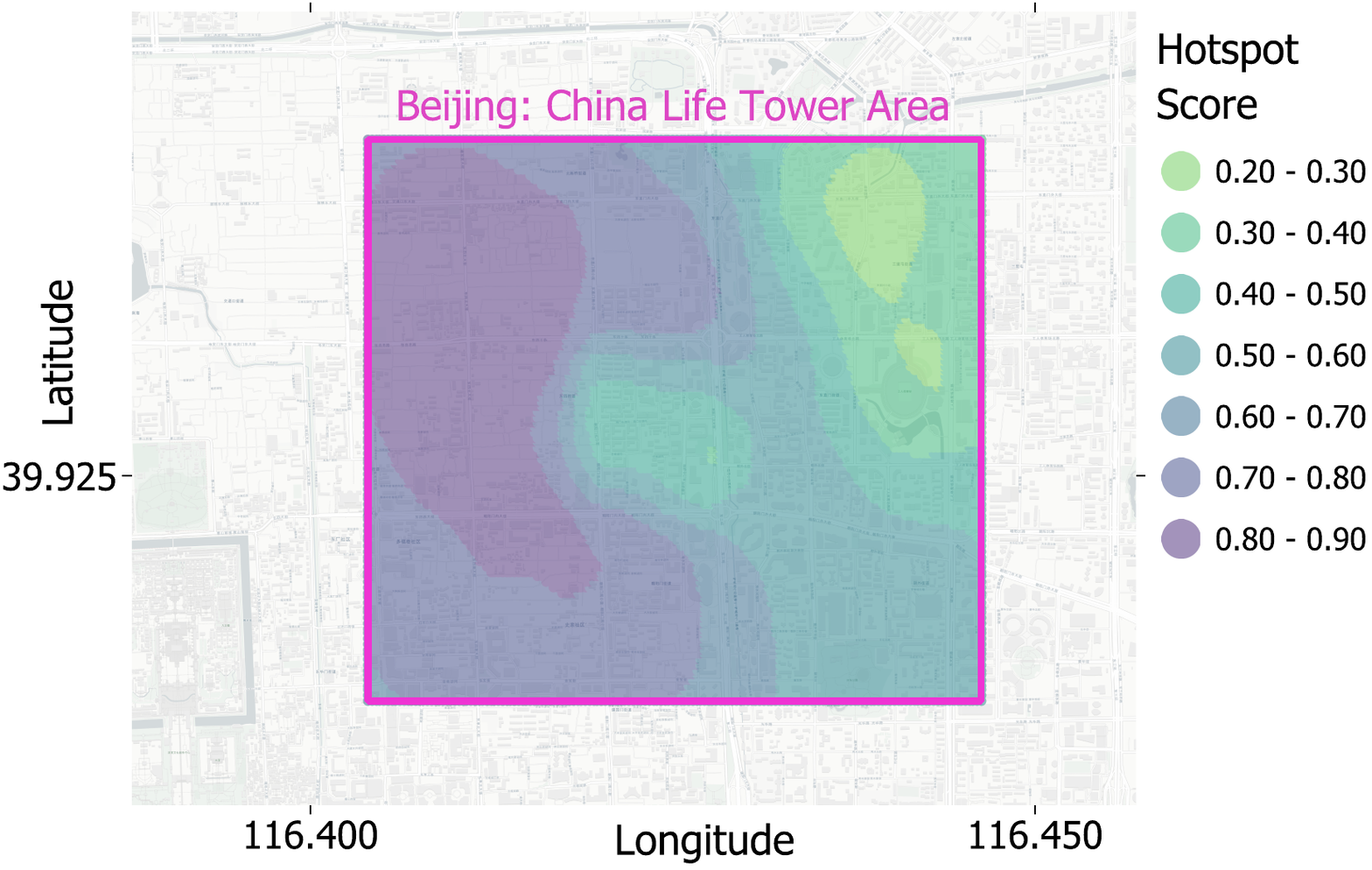}
            \caption{}
            \label{fig: beijing-clt-hotspot-scores}
        \end{subfigure}
    \end{tabular}

    \caption{Spatial $\text{PM}_{2.5}$ distribution from \cite{Wang2023-fa} (a,c) and hotspot scores calculated by our method (b,d).}
    \label{fig: beijing_results_tbl}
\end{figure}

We then applied our hotspot-detection method on the simulated mobile sensing data in the two evaluation areas in Beijing and evaluated the results as described in Section \ref{sec: eval-plots-and-metrics}. The hotspot scores correspond closely with the true spatial $\text{PM}_{2.5}$ profiles. In Beijing LP (Figures \ref{fig: beijing-lp-spatial}, \ref{fig: beijing-lp-hotspot-scores}), the figure-of-eight shape in the Northwest is clearly identifiable, with the mean hotspot score equalling 0.94 among the tiles that constitute this hotspot. The major hotspot along the right and bottom of this quadrant is also evident, with the mean hotspot score equalling 0.98 among the constituent tiles. One also sees that the model delineates the boundaries between high and low pollution, although a caveat is that the estimated boundaries are much smoother than the reality. The boundary between the hotspot on the right and the average pollution region in the middle is only a single tile wide ($30$\si{\metre}). However, the method outputs a boundary seven tiles wide ($210$\si{\metre}). This reflects that the Gaussian process regression surface is overly smooth, and does not effectively model the variations in $\text{PM}_{2.5}$ concentration across small spatial scales. In Beijing CLT (Figures \ref{fig: beijing-clt-spatial}, \ref{fig: beijing-clt-hotspot-scores}), the hotspot scores also match the main spatial contours of $\text{PM}_{2.5}$ pollution. However, the spatial profile of the hotspot scores is again very smooth. In particular, the small patches of varying $\text{PM}_{2.5}$ levels in the left of the quadrant are smoothed over in the hotspot scores.

\begin{figure}[h!]
    \centering
    \includegraphics[width=0.6\textwidth]{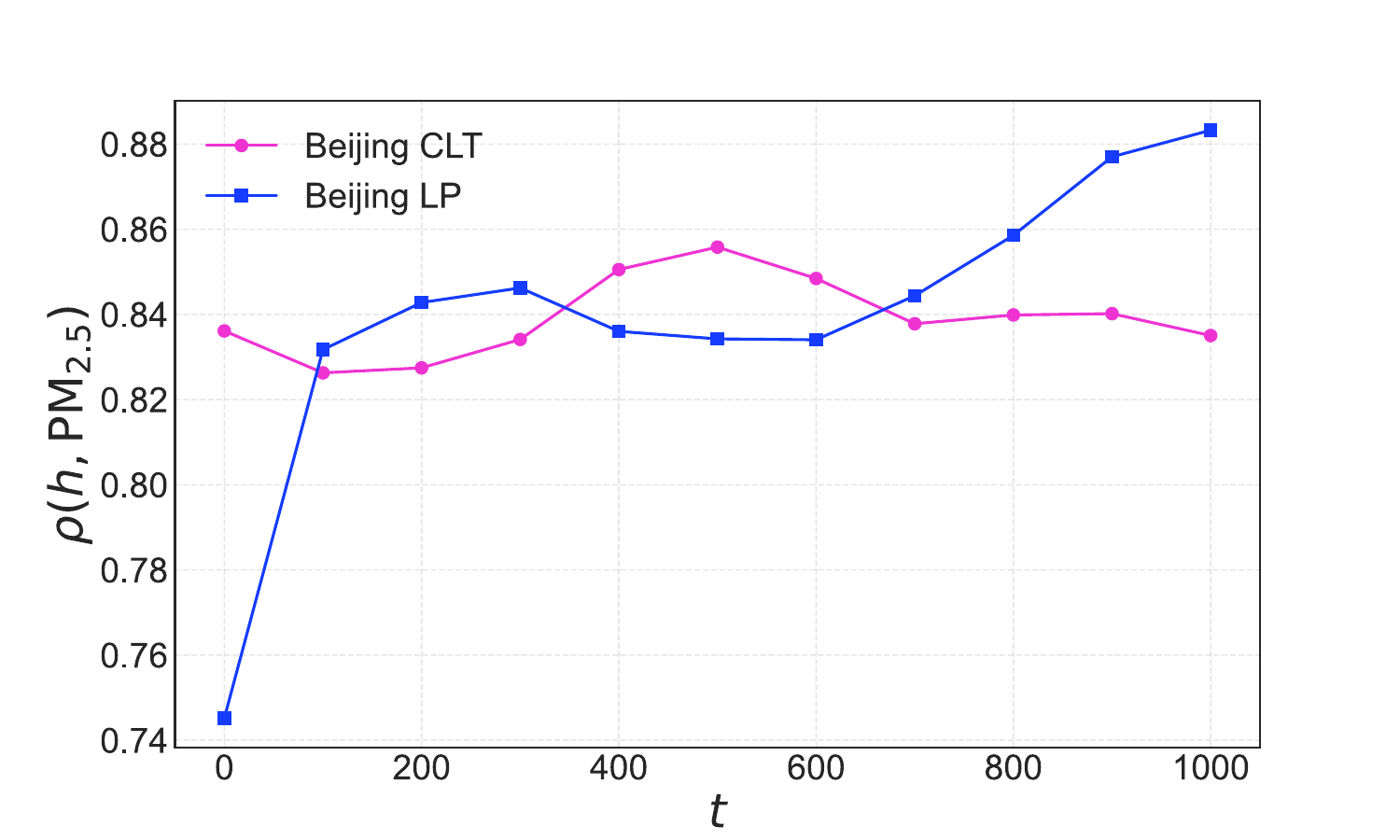}
    \caption{Spearman rank correlation between hotspot scores and spatial $\text{PM}_{2.5}$ averages in Beijing, filtering to tiles with \texttt{n\_measurements} $> t$.}
    \label{fig: beijing-spearman-rank}
\end{figure}

The Spearman rank correlation coefficient value $\rho$ between the hotspot scores and the spatial $\text{PM}_{2.5}$ averages is high in both evaluation areas. When all grid tiles are included, $\rho = 0.8361$ in Beijing CLT and $\rho = 0.7451$ in Beijing LP. These values are close to the maximum achievable correlation of 1, which supports the fact that our method effectively reconstructs the spatial profile of $\text{PM}_{2.5}$ pollution from spatiotemporally sparse mobile sensing data. We also experimented with re-computing the correlation while filtering to the grid tiles where \verb|n_measurements| $> t$ and varying the minimum number $t$, to investigate whether the spatial distribution of the hotspot scores is more accurate among the more frequently sampled grid tiles. Figure \ref{fig: beijing-spearman-rank} shows that as $t$ increases, the correlation increases markedly in Beijing LP but remains stable in Beijing CLT. This suggests that increasing the frequency with which the mobile sensors visit all the tiles could improve the accuracy of the resulting hotspot classification. Future mobile monitoring studies should therefore attempt to ensure that all areas of interest in the
city are visited several times by the drivers carrying the sensors.

Overall, the hotspot scores are reasonably well-calibrated as posterior probabilities (Figure \ref{fig: beijing-calibration}). In both areas, the exceedance proportions in the holdout test set were found to increase monotonically with the hotspot scores estimated on the training set. I.e., the greater the hotspot score in a tile, the more frequently the $\text{PM}_{2.5}$ concentration is indeed exceeding the city-wide median. In Beijing CLT, the calibration is particularly good in the top three bins of hotspot scores, where the calibration curve aligns with the line of perfect calibration. However, for lower bins, the model was generally overconfident: e.g., among tiles where the hotspot score was in the interval (0.3, 0.4], $f(\mathbf{x}_i)$ exceeds $\text{median}(y)$ for only 9.3\% of the observations. In Beijing LP, the calibration curve has an inverse C-shape: the hotspot scores and the observed exceedance proportions are close together at lower and higher values of the hotspot scores, whereas in between the extremities, the calibration curve diverges from the ideal and the model is severely over-confident. Overall, the calibration curves suggest that the hotspot scores are more accurate at extreme hotspots and coldspots, and less accurate in the tiles where the average pollution is close to the city-wide average.

\subsubsection{Calibration with isotonic regression} \label{sec: isotonic}

\begin{figure}[htbp]
  \centering
  \begin{subfigure}[b]{0.45\textwidth}
    \centering
    \includegraphics[width=\textwidth]{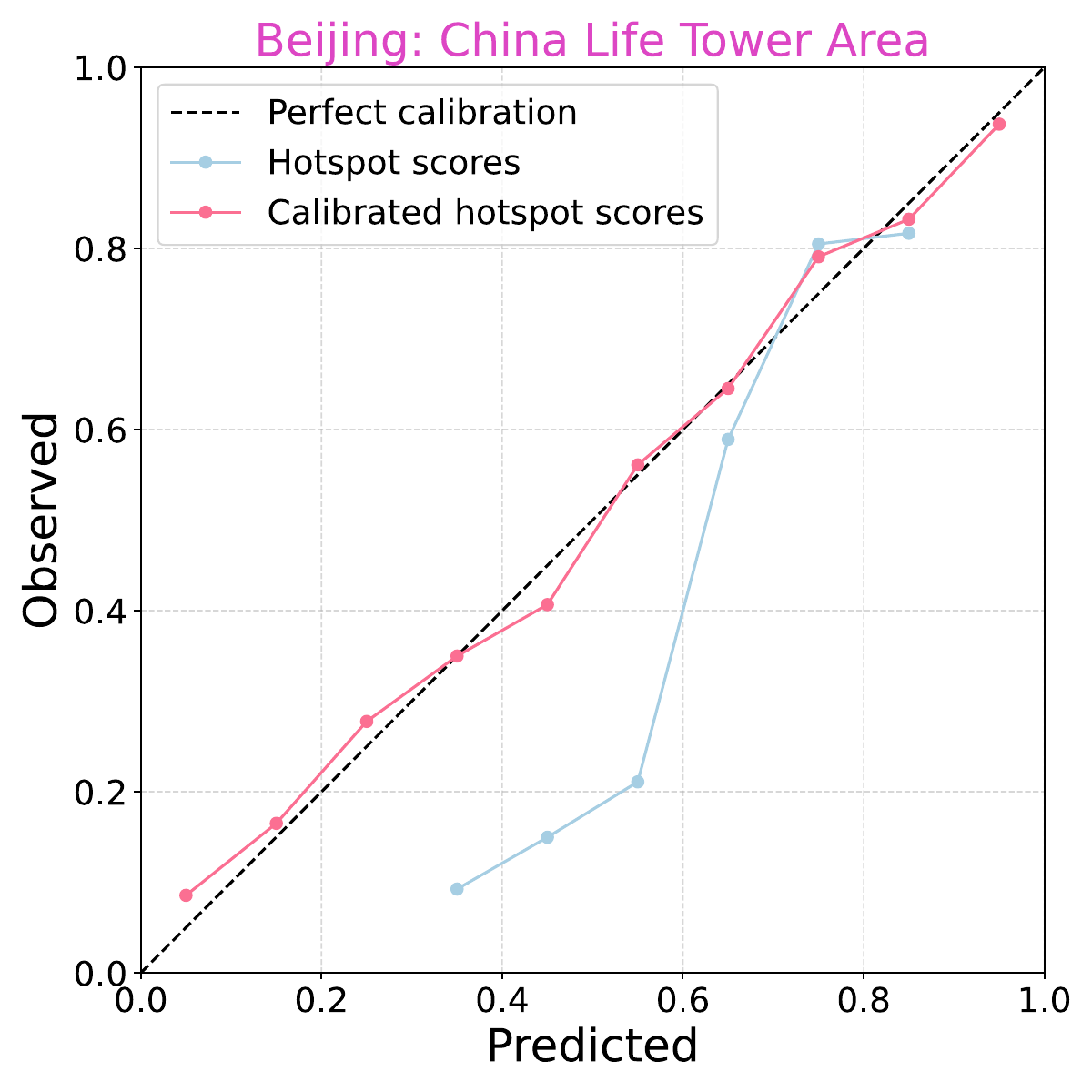}
    \caption{CLT: Reliability diagram}
    \label{fig:clt-reliability}
  \end{subfigure}
  \hfill
  \begin{subfigure}[b]{0.45\textwidth}
    \centering
    \includegraphics[width=\textwidth]{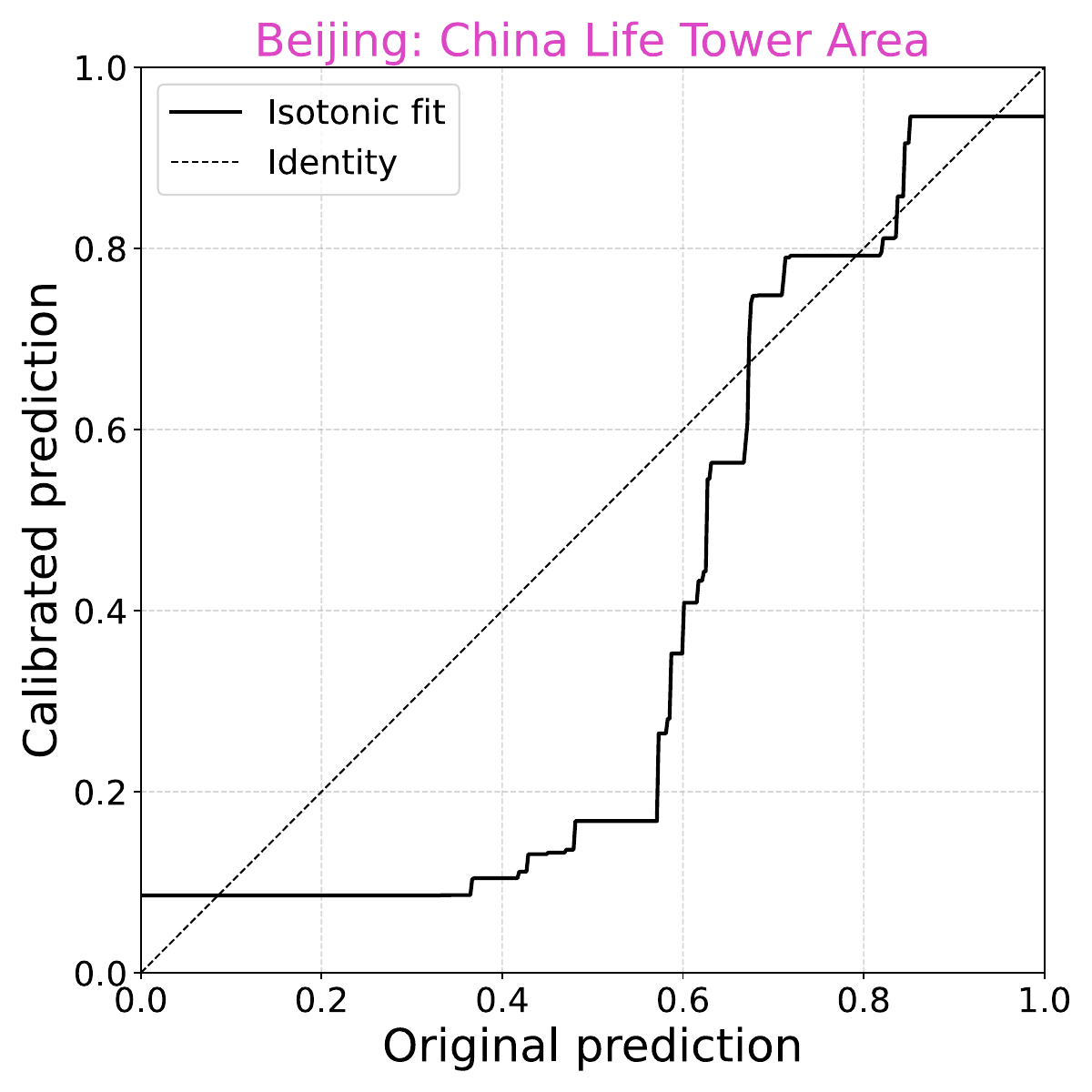}
    \caption{CLT: Isotonic regression fit}
    \label{fig:clt-isotonic}
  \end{subfigure}

  \vskip\baselineskip
  \begin{subfigure}[b]{0.45\textwidth}
    \centering
    \includegraphics[width=\textwidth]{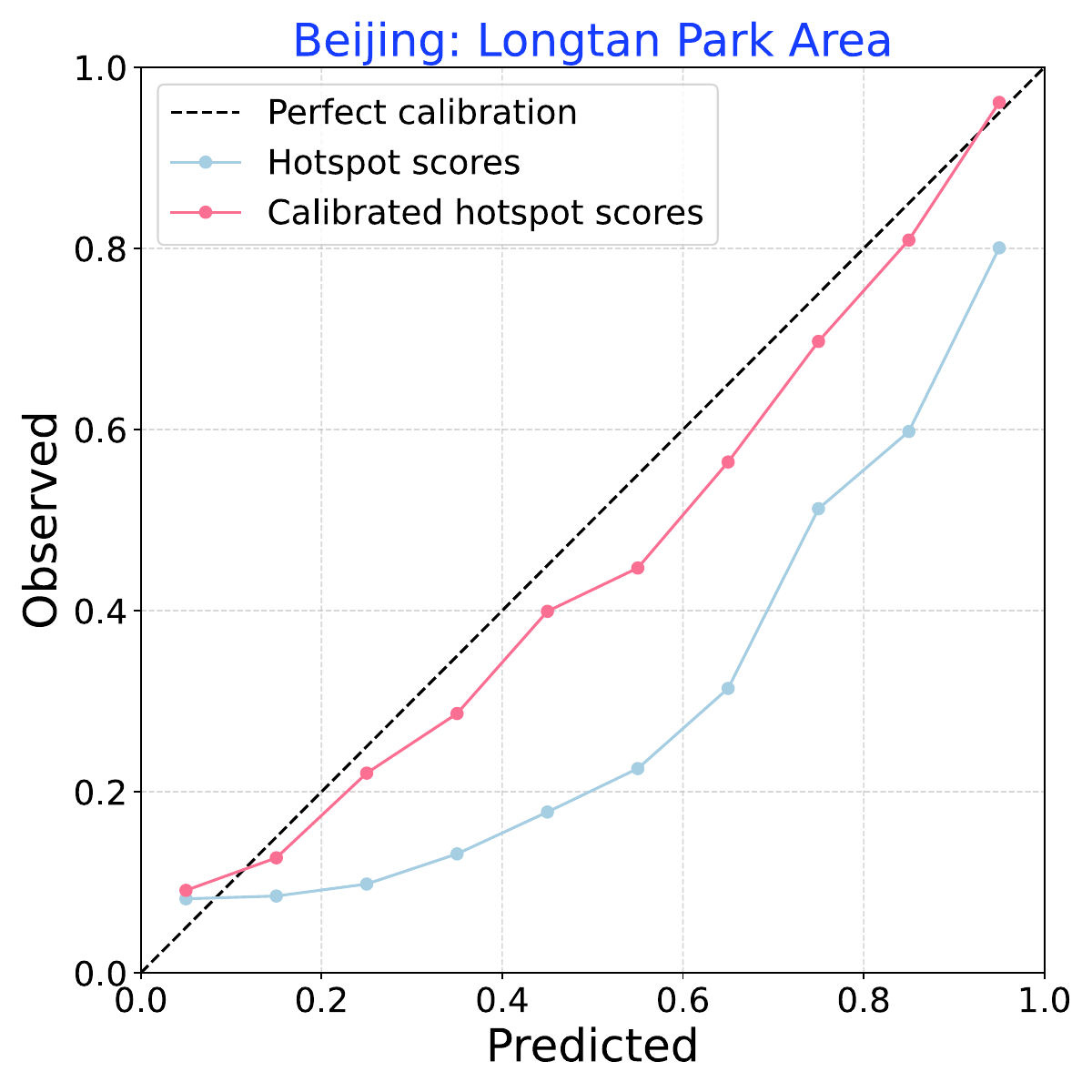}
    \caption{LP: Reliability diagram}
    \label{fig:lp-reliability}
  \end{subfigure}
  \hfill
  \begin{subfigure}[b]{0.45\textwidth}
    \centering
    \includegraphics[width=\textwidth]{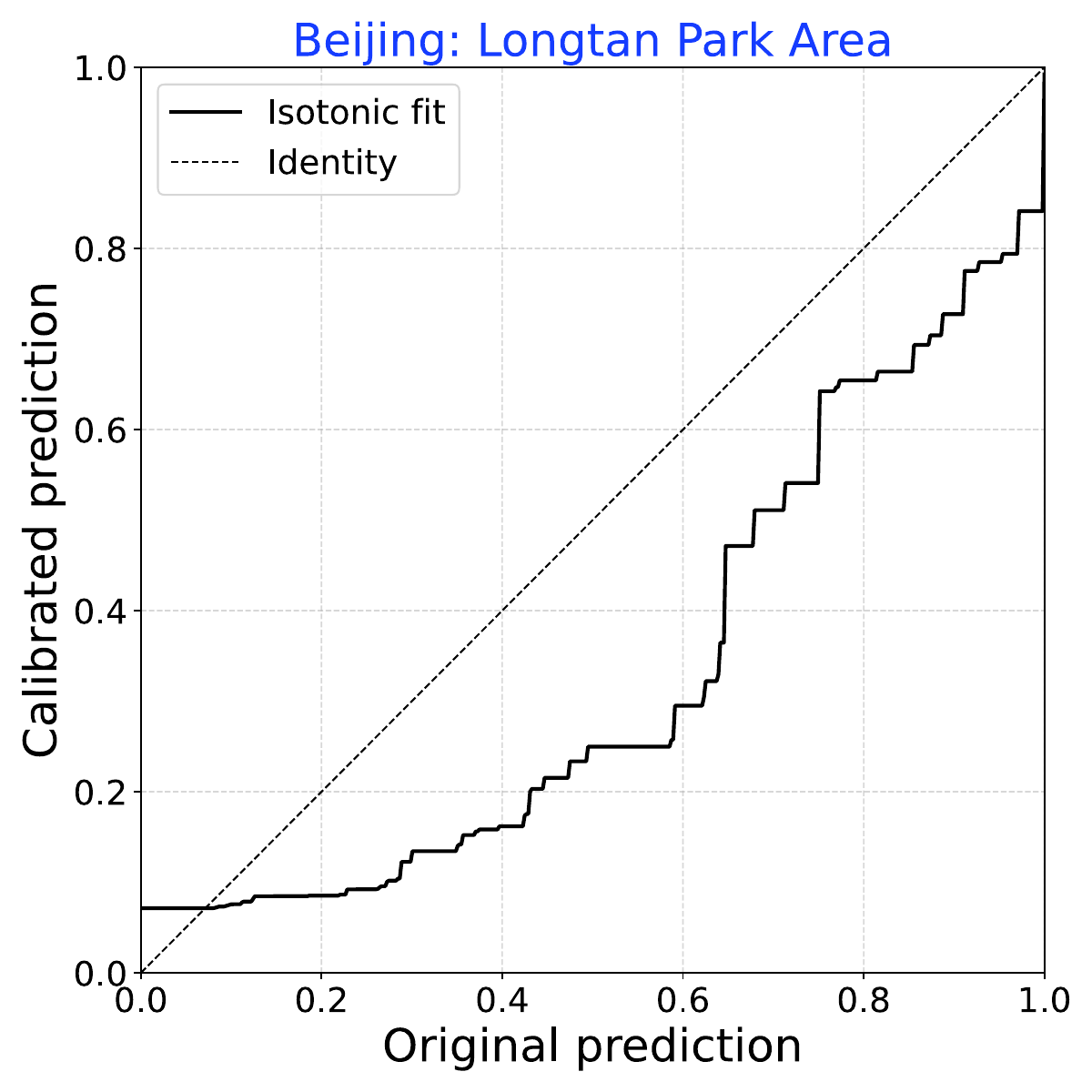}
    \caption{LP: Isotonic regression fit}
    \label{fig:lp-isotonic}
  \end{subfigure}

  \caption{Calibration analysis for the hotspot scores calculated on the Beijing dataset. Reliability diagrams (a,c) compare the predicted hotspot scores against the observed exceedance proportions, before (in blue) and after (in pink) an isotonic regression calibration is applied. Isotonic regression fits (b,d) illustrate how the isotonic regression fit transforms the hotspot scores (x-axis) into calibrated hotspot scores (y-axis).}
  \label{fig: beijing-calibration}
\end{figure}

Figures \ref{fig:clt-isotonic} and \ref{fig:lp-isotonic} illustrate that the main contribution of the isotonic regression models is reducing the hotspot scores estimated by the model for the least-polluted tiles. This is represented by the near-horizontal lines in the bottom left of these two plots. For example, in Beijing CLT (Figure \ref{fig:clt-isotonic}), tiles with a hotspot score up to 0.3307 have a calibrated hotspot score of 0.0855. Similarly in Beijing LP (Figure \ref{fig:lp-isotonic}), tiles with a hotspot score in the range 0.3968 - 0.4228 have a much lower calibrated hotspot score of 0.1618. In Beijing LP (Figure \ref{fig:lp-reliability}), the calibrated hotspot scores are lower than the raw hotspot scores for almost the entire [0,1] interval, reflecting that the model consistently over-estimated the probability of new $\text{PM}_{2.5}$ observations being above average. In Beijing CLT (Figure \ref{fig:clt-reliability}), the calibration regression mostly reduces the hotspot scores too, but in some sub-intervals the isotonic regressor increases the hotspot scores. E.g., tiles with a raw hotspot score between 0.8517-0.9439 have a calibrated hotspot score of 0.9459. Although the method mostly overestimates the exceedance proportions, this highlights that underestimation is also possible.

Calibration with the isotonic regressor greatly improves the key metrics computed on the holdout test set. In Beijing CLT, the ECE decreases from 0.1527 to 0.0081, and the Brier score decreases from 0.2085 to 0.1637. In Beijing LP, calibration reduces the ECE from 0.2294 to 0.0441 and the Brier score from 0.2270 to 0.1688. Figures \ref{fig:clt-reliability} and \ref{fig:lp-reliability} highlight that after transformation with the isotonic regressor, the reliability curve aligns very well with the ideal diagonal curve in both areas.

\FloatBarrier

\section{Discussion}

Short-term exposure to high $\text{PM}_{2.5}$ concentrations is associated with increased PM-related mortality and a wide range of adverse health outcomes \citep{Kloog2013-hq, Yu2024-zt}, hence identifying and cleaning up $\text{PM}_{2.5}$ hotspots is a major opportunity to improve public health, in pursuit of the UN's Sustainable Development Goal 11.6. This study presents a method for identifying $\text{PM}_{2.5}$ hotspots in an urban area, combining mobile sensing with Gaussian process regression. The method consists of four steps: (1) equip a fleet of vehicles with low-cost $\text{PM}_{2.5}$ sensors to measure the air quality; (2) normalise the raw observations for background concentrations by subtracting the median concentration from the previous $x_w$ minutes (in our case, $x_w=15$); (3) fit a Gaussian process regression surface to the normalised observations, and (4) compute posterior probabilities across a spatial grid that the normalised $\text{PM}_{2.5}$ concentration exceeds the city-wide median, which are the `hotspot scores'. Using observed data from Kigali, Rwanda, we showed that our method is compatible with the data from real-world mobile sensing campaigns, and that the hotspots detected by the method intuitively correspond to locations in the city that we would expect to be more polluted. Using simulated mobile sensing data from Beijing, China, we also showed that the hotspot scores preserve the spatial profile of $\text{PM}_{2.5}$ concentrations in a city. I.e., if an area of the city is more polluted than average, then its `hotspot score' is typically higher, and vice versa. We also showed that the hotspot scores are probabilistically well-calibrated, and that the probabilities can be made even more accurate with isotonic regression, an out-of-the-box calibration regressor. These properties suggest that our hotspot scores are apt for identfying $\text{PM}_{2.5}$ hotspots in an urban area, which corroborates the growing enthusiasm for applying Gaussian process regression in spatial statistics problems \citep{Gelfand2016-qt, Christianson2023-rp}. 

Our hotspot-detection method is designed for urban areas in developing countries, where $\text{PM}_{2.5}$ pollution is among the worst in the world and yet fixed $\text{PM}_{2.5}$ monitoring equipment is often lacking \citep{Apte2021-eh}. To this end, only low-cost sensors are used for the data collection, and a background normalisation procedure is proposed that does not require measurements from a fixed $\text{PM}_{2.5}$ monitoring station. Additionally, all computations and analysis were performed with free, open-source software (\verb|R|, \verb|Python|, \verb|QGIS|) and data (\verb|OSM|) so that the method can be recreated without expensive software licences. Furthermore, although Gaussian process modelling is typically computationally intensive, we propose an implementation that greatly reduces the computational requirements of the calculations (Appendix \ref{appendix: gp}). These considerations should remove some of the barriers for practitioners in developing countries to apply our method for their own case study.

In our method, $\text{PM}_{2.5}$ data is collected by equipping a fleet of on-road vehicles with low-cost sensors, then measuring the ambient air quality while the vehicles drive around as usual. We argue that the hotspots identified by our method are therefore highly relevant to the pollution exposure of local road users. The hotspot scores can be paired with raw data about how often each tile was visited in order to assess not only where in the city is the most polluted, but also where road users are most frequently exposed to the bad air quality. There is growing research to suggest that the air quality within cars correlates closely with the outer air quality, particularly when the car windows are open \citep{Kumar2021-oo}. Moreover, motorbike users typically breathe more polluted air than car users because they are less insulated from the outer air \citep{Grana2017-vm}. Accordingly, improving on-road air quality would directly improve the air quality encountered by road users, and our method suggests some priority locations for such an effort.  

A shortcoming of the method is that the estimated Gaussian process regression surface is overly smooth. This is apparent in the Beijing results, where we see that some of the variation in air quality across short distances is smoothed over. However, we found evidence that the hotspot scores better reflect the spatial $\text{PM}_{2.5}$ pattern when filtering out the least-sampled tiles, which suggests that one way to increase the spatial resolution of the Gaussian process model regression is to ensure that all tiles are well-sampled. E.g., by conducting a longer mobile sensing campaign than one month, or by using a larger number of low-cost sensors. Another avenue to explore for improving the spatial resolution of the model is experimenting with the parameters of the Gaussian process regression model. E.g., future research could explore a larger variety of kernel choices.

A separate shortcoming is that the simulation process for the synthetic mobile sensing data in Beijing is overly simplified. Synthetic $\text{PM}_{2.5}$ measurements were generated according to Equation \ref{eq: synthetic-pm25}, considering the spatial $\text{PM}_{2.5}$ distribution in Beijing, the hourly trends in real-world monitoring station data, and two sources of stochastic measurement interference. State-of-the-art air pollution simulations generally incorporate atmospheric dispersion, explicit pollution sources, meteorological conditions, and sometimes traffic patterns \citep{EPA_AirQualityModels, Baek2022-yw}. However, the main purpose of the mobile sensing data simulation is to generate data with a high spatial and temporal resolution where the spatial $\text{PM}_{2.5}$ profile is known, so that we can assess the accuracy of our method with reference to a ground truth. The realism of the air pollution dynamics is therefore of secondary importance. Nevertheless, a mobile sensing simulator with high spatiotemporal resolution \emph{and} a greater incorporation of real-world air pollution dynamics would enable a more convincing evaluation of the method.

In Section \ref{sec: isotonic} we showed that calibrating the hotspot scores with an isotonic regression fit makes the posterior probability estimates more accurate. Therefore, an interesting direction for future work is to devise a calibration procedure that also works on real, non-simulated mobile sensing data. The calibration procedure here relies on the fact that noise-free observations of $f(\mathbf{x})$ are available when the dataset is simulated, but this is not the case in a real-world mobile sensing campaign, where observations $y_i$ are theorised to be corrupted by some random noise. Future studies could consider combining the measurements from multiple sensors or triangulating the mobile measurements with a reference station in order to estimate $f(\mathbf{x})$, and then using the estimates $\hat{f}(\mathbf{x})$ to calibrate the hotspot scores. Nevertheless, we reiterate that the calibration transformation does not change the relative ranking of the grid tiles, so the hotspot scores can still be useful to identify hotspots even if poorly calibrated. 

Future research could also incorporate meteorological conditions into the method. Relative humidity (RH) is known to affect the accuracy of low-cost particle sensors \citep{Crilley2018-sn, Zou2021-td}. Adjusting the raw $\text{PM}_{2.5}$ observations based on the RH at measurement time might therefore improve the accuracy of the $\text{PM}_{2.5}$ measurements, making the resulting hotspot classification more accurate. Wind dynamics can also affect the urban air quality, with changes in wind pattern typically accompanied by a short-term deterioration in air quality \citep{Xie2022-ep}. Wind speed or wind direction could be included as predictors in the Gaussian process regression model to further refine the accuracy of the estimated spatial $\text{PM}_{2.5}$ profile. 

Finally, the method can also be adapted to answer questions about how the pollution hotspots vary by time of day, since diurnal $\text{PM}_{2.5}$ patterns are of great interest \citep{Ye2018, Ramachandran2000, Du2020}. A natural approach would be to divide the data into time brackets, compute the hotspot scores for each bracket, and then compare the spatial profile of hotspot scores at different times of day. This would only require checking that there is sufficient data in each of the time brackets, since our mobile sensing approach generally results in peak hours with many observations and quiet hours with few observations.

\section{Conclusion}

This study introduced a method for detecting $\text{PM}_{2.5}$ hotspots in an urban area using mobile sensing and Gaussian process regression. The method outputs spatially-explicit `hotspot scores' corresponding to the estimated probability that a $\text{PM}_{2.5}$ observation exceeds the city-wide median, while controlling for the background ambient pollution and excluding the random noise variable. The hotspot scores provide a convenient overview of the relative pollution levels throughout the city. Moreover, the method is tailored to urban areas in developing countries, where $\text{PM}_{2.5}$ pollution is among the worst in the world yet $\text{PM}_{2.5}$-monitoring infrastructure is often lacking. To this end, the only resources required to re-apply the method on a new case study are low-cost mobile sensors and free, open-source software.

The method was evaluated using observed mobile sensing data in Kigali, Rwanda. We found that particulate matter pollution in Kigali consistently exceeds the safe limits for human health established by the World Health Organisation, highlighting the need for local action to improve the air quality. The hotspots proposed by our method in Kigali correspond to intuitive urban hotspot locations: busy roads and intersections in the city centre, which supports the validity of the hotspot scores. The method was also evaluated using spatial average $\text{PM}_{2.5}$ concentration data from Beijing, China. Here it was demonstrated that the proposed hotspot scores effectively discern the spatial distribution of $\text{PM}_{2.5}$ in an urban area. Additionally, it was shown that the hotspot scores are probabilistically well-calibrated, and that the calibration can be further improved using an out-of-the-box Isotonic regression procedure. 


Our method can be re-applied in cities throughout the world to help fill the gap in urban air quality information and empower public health officials. A key direction for future work will be re-applying the method in different cities, both to continue increasing the availability of air quality data and to ensure that the method works for urban areas all over the world. Other priorities for future work include improving the spatial resolution of the Gaussian process model predictions, devising a probabilistic calibration procedure that works with real-world mobile sensing data, and incorporating meteorological data into the method. 

\section*{Acknowledgements}

We would like to thank Krzysztof Cybulski and Lukas Graz from the Statistical Consulting team at ETH Zürich for their advice on the methodology. We also thank the following partners for funding the data collection in Kigali: the Wuppertal Institute, UN-Habitat, and the United Nations Environment Programme (UNEP), and the International Climate Protection Initiative (IKI) of the German Ministry of the Environment. Additionally, we thank the following partners who assisted with collecting the data: the University of Rwanda, the City of Kigali, the Urban Electric Mobility Initiative, and Ampersand Rwanda Ltd. Those affiliated with the University of Cambridge acknowledges funding by the Engineering and Physical Sciences Research Council Centre for Doctoral Training in Sensor Technologies and Applications (EP/L015889/1). We thank the Centre for Global Equality (CGE) for its continued support of open-seneca and its involvement in the Urban Pathways project. We thank Yongyue Wang from Tsinghua University for kindly sharing the Beijing data.

\section*{Code availability}

The underlying code for this study is not publicly available but may be made available to qualified researchers on reasonable request from the corresponding author.

\section*{Data availability}

The collected and processed Kigali data is available at https://zenodo.org/records/15206350. 

\section*{Declaration of competing interests}

The authors declare that they have no known competing financial
interests or personal relationships that could have appeared to influence
the work reported in this paper.

\section*{Funding sources}

This work was supported with sensors from open-seneca. Additionally, the following partners helped finance the data collection in Kigali:
the Wuppertal Institute, UN-Habitat, and the United Nations Environment Programme (UNEP), and the International Climate Protection Initiative (IKI) of the German Ministry of the Environment. 


\appendix

\newpage

\section{Computing hotspot scores with Gaussian Process Regression} \label{appendix: gp}

This Appendix elaborates on how we apply Gaussian process regression to compute the tilewise hotspot scores in step 4 of our hotspot-detection methodology. We first introduce Gaussian processes from probability and the Gaussian process regression setup. Then we explain how we fit the Gaussian process regression to our data, before elaborating on how the hotspot scores are computed and their interpretation.

\subsubsection{Gaussian processes}

A Gaussian process is a stochastic process - i.e., an infinite collection of random variables $\{X_t\}$ indexed by some $t$ (usually time) - in which any finite subset of random variables have a multivariate Gaussian distribution. I.e., $\forall A = \{X_1, \ldots, X_n\} \subseteq \{X_t\}, \ A \sim N(\mathbf{\mu}_A, \mathbf{\Sigma}_{AA})$.

\subsubsection{Gaussian process regression}

As in the general regression setting, the objective in Gaussian process regression is to learn a function $f$ relating a quantity of interest $y_i$ to a known quantity $\mathbf{x}_i$ in the presence of random noise $\varepsilon_i$:

\[ y_i = f(\mathbf{x}_i) + \varepsilon_i \]

In Gaussian process regression, it is additionally assumed that $f$ has a Gaussian process as prior distribution.  I.e., that for all subsets of the input domain $A = \{\mathbf{x}_1, ..., \mathbf{x}_n\} \subset \mathcal{X}$, $\mathbf{f}_A \sim \mathcal{N}(\mu_{ \mathbf{f}_A},\mathbf{K}_A)$, where $\mathbf{f}_A = [f(\mathbf{x}_1), ... , f(\mathbf{x}_n)]^\top$. $\mu_{ \mathbf{f}_A} = [ \mu_1, ... , \mu_n]^\top$ is the vector of mean values of the function at $\mathbf{x}_i \ (i = 1, \ldots, n)$ and $\mathbf{K}_A \in \mathbb{R}^{n\times n}$ is the covariance matrix of $\mathbf{f}_A$ with elements $\mathbf{K}_{i,j} = k(\mathbf{x}_i, \mathbf{x}_j)$. $k: \mathcal{X} \times \mathcal{X} \rightarrow \mathbb{R}$ is the covariance function of the Gaussian process prior, selected according to one's beliefs about the dependency between values of the function $f$ evaluated at nearby points $x_i$, $x_j$. Furthermore, $\varepsilon_i$ are typically assumed to be i.i.d N(0, $\sigma_{ \varepsilon}^2$) random variables \citep{rasmussen_2005-sv}. 

\subsubsection{Gaussian Process regression applied to $\text{PM}_{2.5}$ data}

In our context, $i$ indexes an observation from the data. $\mathbf{x}_i \in \mathcal{X} \subset \mathbb{R}^2$ is the vector of latitude and longitude coordinates at $i$, $y_i$ denotes the normalised $\text{PM}_{2.5}$, and $f$ is the random variable relating a pair of coordinates to its $\text{PM}_{2.5}$ concentration. We argue that Gaussian process regression is an apt choice for modelling $\text{PM}_{2.5}$ concentrations because at any given (lat, lon) location, Gaussian process regression represents the spatial $\text{PM}_{2.5}$ concentration as a \emph{probability distribution} as opposed to a single value. Therefore, the method accounts for the remaining uncertainty in the $\text{PM}_{2.5}$ concentration which is not explained by the background pollution at the time of the measurement or by random noise.

Throughout this work we apply the exponential kernel as the covariance function $k$:

\[ k( \mathbf{x}_i,\mathbf{x}_j) = \sigma_k^2\exp\left(- \ \frac{\lVert \mathbf{x}_i - \mathbf{x}_j\rVert_1}{ \ell }\right) \]

It is a `spiky' kernel which captures the fact that $\text{PM}_{2.5}$ concentrations exhibit a high variation across short geographic distances. $\ell$ is the length-scale parameter which models the `spikiness' of the kernel, and $\sigma^2_k$ is the kernel variance parameter. Both parameters are optimised during model training by \verb|gpflow| \citep{Matthews2016-ca}. Other researchers have typically preferred the squared exponential kernel for modelling $\text{PM}_{2.5}$ concentrations with Gaussian process regression \citep{Cheng2014-fm, Stoddart2023-ug, Patel2022-rv}. However, empirical tests on the Kigali dataset suggested that the exponential kernel performs better at hotspot detection. 

\subsubsection{Bagging Gaussian process fits}

As described by \cite{rasmussen_2005-sv}, `One issue with Gaussian process prediction methods is that their basic complexity is $\mathcal{O}(n^3)$, due to the inversion of a $n \times n$ matrix. For large datasets this is prohibitive (in both time and space)'. With mobile sensing datasets typically comprising a large number of observations - e.g., we have $n > 2$ million observations in the processed Kigali dataset - fitting a single Gaussian process to all of the mobile sensing observations can be computationally infeasible. Hence we instead apply simple bagging with subsamples of the data \citep{Breiman1996-tz}, as detailed in the Gaussian process context by \cite{Chen2009-uw}. Instead of estimating one spatial $\text{PM}_{2.5}$ distribution for the city extent, we take $B$ random samples of size $m << n$ from the mobile sensing data and fit a separate model $\hat{f}_b$ to each sample $b = 1, \ldots, B$. Predicted means and variances at a point $\mathbf{x}_i$ are computed by averaging the $B$ individual predictions according to:

\begin{align}
    \hat{\mu}(\mathbf{x}_i) &= \frac{1}{B} \sum_{b=1}^B \hat{ \mu_b}(\mathbf{x}_i) \label{eq: bag mean} \\
    \hat{\sigma}^2 (\mathbf{x}_i) &= \frac{1}{B} \sum_{b=1}^B \hat{\sigma_b}^2(\mathbf{x}_i) + \frac{1}{B} \sum_{b=1}^B (\hat{ \mu_b}(\mathbf{x}_i) - \hat{\mu}(\mathbf{x}_i))^2 \label{eq: bag var}
\end{align}

where $\hat{ \mu_b}(\mathbf{x}_i)$ and $\hat{\sigma_b}^2(\mathbf{x}_i)$ are respectively the mean and variance values of f($\mathbf{x}_i$) predicted by the $b$'th model. However, unlike \cite{Chen2009-uw}, we do not select observations for the $B$ samples uniformly at random. Instead, we employ a scheme to disproportionately sample from the most polluted and least polluted grid tiles, to ensure that more data is gathered about tiles suspected to be hotspots or coldspots. Our sampling scheme orders the grid tiles according to their median normalised $\text{PM}_{2.5}$ concentration. Then, we specify probabilities $p_{\text{high}}$ and $p_{\text{low}}$ and thresholds $r_{\text{high}}$ and $r_{\text{low}}$, and proceed to sample observations from the top $r_{\text{high}}$ proportion of tiles with probability $p_{\text{high}}$ and from the bottom $r_{\text{low}}$ proportion of tiles with a probability proportional to $p_{\text{low}}$. In our experiments we chose $p_{\text{high}} = p_{\text{low}} = 0.3$ and $r_{\text{high}} = r_{\text{low}} = 0.2$, meaning observations belonging to the most fifth and the least polluted fifth of tiles are 50\% more likely to appear in a bootstrapped sample than the rest. 

All experiments were run on a machine with a 12th Gen Intel(R) Core i7-1260P CPU (12 cores, 16 processors), 16GB RAM, running Microsoft Windows 11. The code was executed using \verb|Python 3.11.7| with version 2.9.1 of package \verb|gpflow|. We bagged $B = 100$ models fit to subsets of $m = 2000$ datapoints in the experiments. The process of fitting the models, predicting the mean and variance across the spatial grid and then combining the results typically took around 3 hours to execute.

\subsubsection{Calculating hotspot scores}

Once the $B$ gaussian process regression surfaces have been fit, we predict the mean and variance of $f$ at every tile centroid $\mathbf{x}_j$ in the grid according to Equations \ref{eq: bag mean} and \ref{eq: bag var}. By the marginalisation property of Gaussian processes \citep{rasmussen_2005-sv}, the normalised $\text{PM}_{2.5}$ $f(\mathbf{x}_j)$ at a tile centroid $\mathbf{x}_j$ is a random variable with the distribution $N(\mu(\mathbf{x}_j), \sigma^2( \mathbf{x}_j))$ which we estimate by $N(\hat{\mu}(\mathbf{x}_j), \hat{\sigma}^2( \mathbf{x}_j))$. Denoting by $\text{median}(y)$ the estimated median of normalised $\text{PM}_{2.5}$ throughout the city, the estimated probability that $f( \mathbf{x}_j)$ exceeds the city-wide median $\text{PM}_{2.5}$:

\[  \hat{\mathbb{P}}[ f( \mathbf{x}_j) > \text{median}(y) ] \]

thus equals:

\[ \mathbb{P}[ Z > \frac{\text{median}(y) - \hat{\mu}(\mathbf{x}_j)}{\hat{\sigma}( \mathbf{x}_j)} ] \]

where $Z \sim N(0,1)$ is a standard Gaussian variable. This probability is exactly the hotspot score reported in this study.

One can compare the hotspot scores against a desired critical value $p_{\text{crit}}$ in order to perform a binary classification of which tiles are hotspots and which are not. By setting $p_{\text{crit}} = 0.95$ for example, we consider as hotspots the tiles where we estimate that with a probability of 95\%, an observation of the normalised $\text{PM}_{2.5}$ value will exceed the median normalised $\text{PM}_{2.5}$ in the city.

\subsubsection{Interpretation of the hotspot scores}

The interpretation of the hotspot scores is also nuanced. In particular, we stress that our hotspot scores are different from p-values, including from the Bayesian variant posterior predictive p-values conceptualised by \cite{Rubin1984-wh} and \cite{Meng1994-og}. The confidence score at a (lat, lon) location is the posterior probability that a normalised $\text{PM}_{2.5}$ measurement will exceed the (empirical) normalised city-wide median. By contrast, p-values measure the discrepancy between observed data and some conjecture about an underlying probability distribution. P-values belong to the toolkit of statistical hypothesis testing, whereas our method does not test any statistical hypothesis. 


\newpage

\section{Co-location with reference station, Stockholm, Sweden} \label{appendix: co-location}

\begin{figure}[htbp]
  \centering
  \begin{subfigure}[b]{0.45\textwidth}
    \centering
    \includegraphics[width=\textwidth]{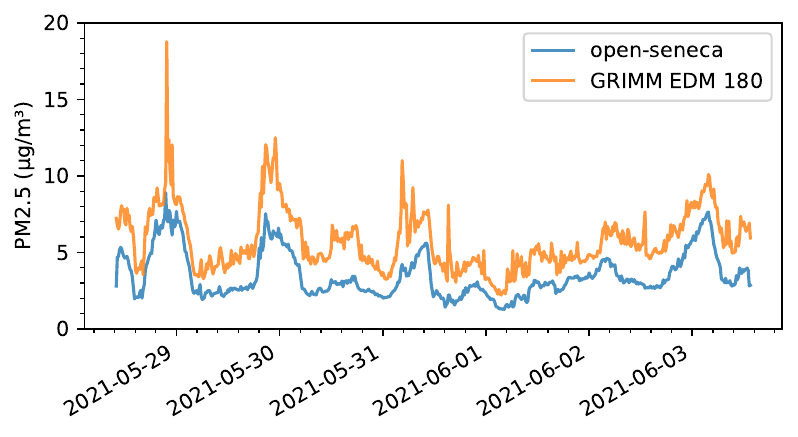}
    \caption{}
    \label{fig:co-location-time-series}
  \end{subfigure}
  \hfill
  \begin{subfigure}[b]{0.45\textwidth}
    \centering
    \includegraphics[width=\textwidth]{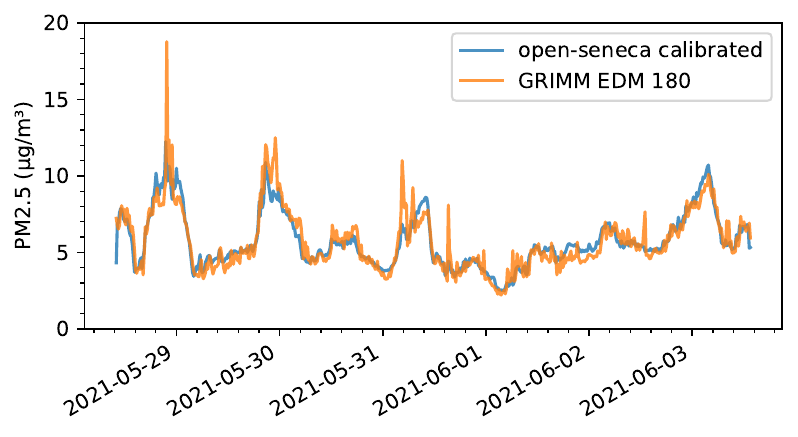}
    \caption{}
    \label{fig:co-location-time-series-post-calibration}
  \end{subfigure}

  \vskip\baselineskip
  \begin{subfigure}[b]{0.45\textwidth}
    \centering
    \includegraphics[width=\textwidth]{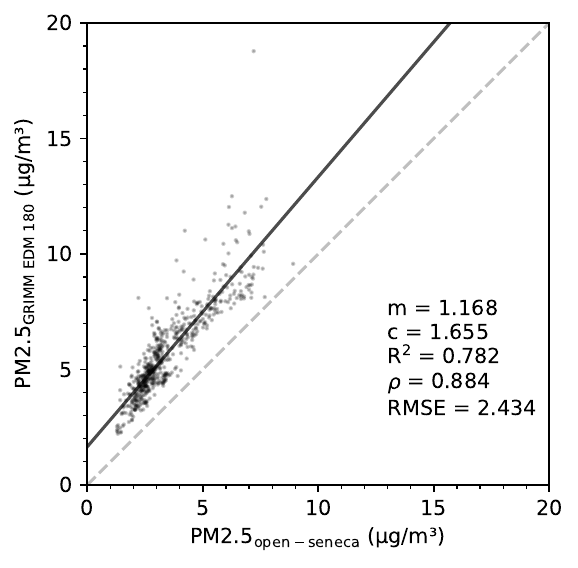}
    \caption{}
    \label{fig:co-location-scatter}
  \end{subfigure}
  \hfill
  \begin{subfigure}[b]{0.45\textwidth}
    \centering
    \includegraphics[width=\textwidth]{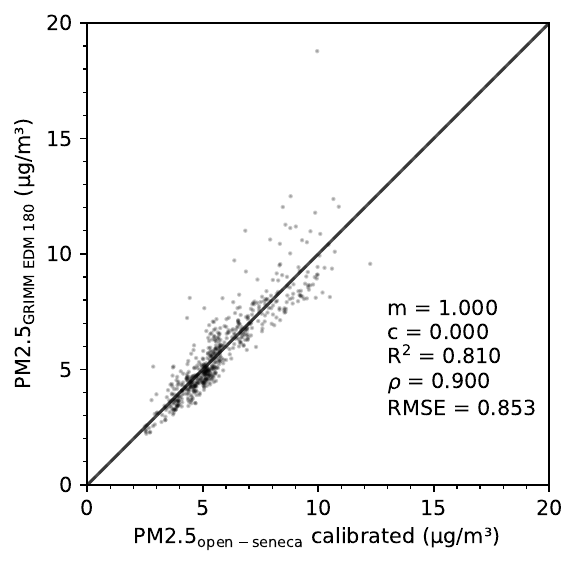}
    \caption{}
    \label{fig:co-location-scatter-post-calibration}
  \end{subfigure}

  \vskip\baselineskip
  \begin{subfigure}[b]{0.45\textwidth}
    \centering
    \includegraphics[width=\textwidth]{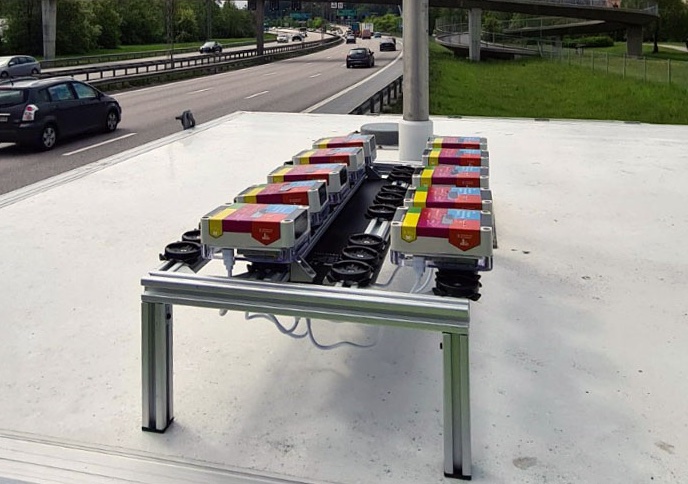}
    \caption{}
    \label{fig:co-location-image}
  \end{subfigure}

  \caption{Co-location of 10 open-seneca sensors with a GRIMM EDM 180 reference station next to a motorway near Stockholm, Sweden (GPS coordinates [59.2475, 17.8432]) over a one week period. Panels (a) and (b) show time-series median measurements from the open-seneca sensors compared to the reference before and after a linear calibration. Panels (c) and (d) display corresponding PM2.5 behavior with linear fits, where m is the slope, c the intercept, R$^2$ the coefficient of determination, $\rho$ the Pearson correlation coefficient, and RMSE the root-mean-squared error. Panel (e) shows the 10 open-seneca sensors mounted on the reference station. During the co-location, the relative humidity varied between 29.9\% and 98.0\%, where some minor PM2.5 discrepancies from reference were likely due to high-humidity changing the particle size distribution measured by open-seneca's optical particle counter.}
  \label{fig: os-co-location}
\end{figure}

\newpage

\newpage


\end{document}